%% file: manuscript.tex
\documentclass[a4paper]{llncs}  
\usepackage{hyperref}

\usepackage{amsmath,amssymb}
\usepackage{mathtools}

\usepackage{algorithm}
\usepackage[noend]{algpseudocode}

\usepackage{booktabs}
\usepackage{multirow}
\usepackage{wrapfig}

\usepackage{xcolor}

\newcommand{\Reals}{\mathbb{R}}

\newcommand{\Set}[1]{\left\{#1\right\}}
\newcommand{\Tuple}[1]{\left(#1\right)}
\newcommand{\Range}[2]{\left[#1,\ #2\right]}
\newcommand{\ORange}[2]{\left(#1,\ #2\right)}
\newcommand{\LOrange}[2]{\left(#1,\ #2\right]}

\newcommand{\Abs}[1]{{\left|#1\right|}}

\newcommand{\Iverson}[1]{\left[#1\right]}

\newcommand{\Fn}[2]{#1\left(#2\right)}
\newcommand{\FnApp}[2]{\Fn{#1}{#2}}
\newcommand{\FnDef}[3]{#1: #2 \rightarrow #3}

\newcommand{\Exp}[1]{e^{#1}}

\newcommand{\ProbSymbol}{P}
\newcommand{\Prob}[2][]{\FnApp{\ProbSymbol_{#1}}{#2}}

\newcommand{\DS}{\mathcal{D}}
\newcommand{\PattLang}{\mathcal{L}}
\newcommand{\PattSet}{\mathcal{P}}
\newcommand{\Items}{\mathcal{I}}
\newcommand{\Transactions}{\mathcal{T}}
\newcommand{\Item}{i}
\newcommand{\Transaction}{t}
\newcommand{\Pattern}{p}
\newcommand{\Constraints}{\mathcal{C}}
\newcommand{\MinSup}{\theta}

\newcommand{\Qual}{\varphi}
\newcommand{\FreqW}{freq}
\newcommand{\Freq}[2][]{\FnApp{\FreqW{}^{#1}}{#2}}
\newcommand{\ChiSq}{\chi^2}
\newcommand{\SurpW}{surp}

\newcommand{\Ranking}{R}
\newcommand{\Hyp}{h}

\newcommand{\Feedback}{U}

\newcommand{\Coeff}{w}

\newcommand{\Formula}{\mathbf{F}}
\newcommand{\Assignment}{F}

\newcommand{\Weight}{w}
\newcommand{\WFn}[1]{\FnApp{\Weight}{#1}}

\newcommand{\WgCellSmp}{\varsigma}

\newcommand{\QualLog}{\Qual_{\text{logistic}}}
\newcommand{\Query}{Q}
\newcommand{\DesiredTilt}{A}
\newcommand{\QuerySize}{k}
\newcommand{\QueryOverlap}{l}

\newcommand{\MeanSD}[2]{$#1 \pm #2$}
\newcommand{\MeanSDBest}[2]{$\mathbf{#1 \pm #2}$}

\newcommand{\Ptn}{\Pattern}
\newcommand{\Q}{\Query}
\newcommand{\A}{\DesiredTilt}
\newcommand{\QS}{\QuerySize}
\newcommand{\QO}{\QueryOverlap}

\newcommand{\LongShort}[2]{%
	\ifdefined\LONGVER
		#1%
	\else
		#2%
	\fi
}

\newcommand{\LongOnly}[1]{%
	\LongShort{#1}{}%
}

\newcommand{\ShortOnly}[1]{%
	\LongShort{}{#1}%
}

\newcommand{\Algo}[1]{\textsc{#1}}
\newcommand{\ParHead}[1]{\smallskip

\noindent \textbf{#1}}

\newcommand{\HPh}{\hphantom{1}}

\newcommand{\Rows}[2]{\multirow{#1}{*}{#2}}
\newcommand{\ColC}[2]{\multicolumn{#1}{c}{#2}}

\newcommand{\Wg}{\Algo{WeightGen}}

\newcommand{\Flexics}{\Algo{Flexics}}
\newcommand{\AlgName}{\Algo{LetSIP}}
\newcommand{\Aples}{\Algo{APLe}}
\newcommand{\Bah}{\Algo{IPM}}
\newcommand{\Prime}{\Algo{Priime}}

\newcommand\blankfootnote[1]{%
  \begingroup
  \renewcommand\thefootnote{}\footnote{#1}%
  \addtocounter{footnote}{-1}%
  \endgroup
}

	\newcommand*{\LONGVER}{}
\makeatletter%
\begin{document}

\title{Learning what matters -- \\ Sampling interesting patterns}
\author{%
	Vladimir Dzyuba\inst{1}
		\and
	Matthijs van Leeuwen\inst{2}%
}
\institute{%
	Department of Computer Science, KU Leuven, Belgium
		\and
	LIACS, Leiden University, The Netherlands\\
	\email{vladimir.dzyuba@cs.kuleuven.be}, 
	\email{m.van.leeuwen@liacs.leidenuniv.nl}%
}
\date{}

\maketitle
\begin{abstract}
In the field of \emph{exploratory data mining}, local structure in data can be described by \emph{patterns} and discovered by mining algorithms. Although many solutions have been proposed to address the redundancy problems in pattern mining, most of them either provide succinct pattern sets or take the interests of the user into account---but not both. Consequently, the analyst has to invest substantial effort in identifying those patterns that are relevant to her specific interests and goals.

To address this problem, we propose a novel approach that combines pattern sampling with interactive data mining. In particular, we introduce the \AlgName{} algorithm, %
which builds upon recent advances in 1) weighted sampling in SAT and 2) learning to rank in interactive pattern mining. Specifically, it \emph{exploits user feedback to directly learn the parameters of the sampling distribution that represents the user's interests}.

We compare the performance of the proposed algorithm  to the state-of-the-art in interactive pattern mining by emulating the interests of a user. The resulting system allows efficient and interleaved learning and sampling, thus user-specific anytime data exploration. Finally, \AlgName{} demonstrates favourable trade-offs concerning both quality--diversity and exploitation--exploration when compared to existing methods.
\end{abstract}

\section{Introduction}
\label{sec:introduction}

\LongOnly{\blankfootnote{
This document is
an extended version of
a conference publication
\cite{Dzyuba2017b}.}}

Imagine a data analyst who has access to
a medical database containing information about
patients, diagnoses, and treatments.
Her goal is to identify
novel connections between
patient characteristics and
treatment effects.
For example, one treatment may be
more effective than another 
for patients of a certain age and occupation,
even though the latter is
more effective at large.
Here, age and occupation are
latent factors that
explain the difference in
treatment effect.

In the field of \emph{exploratory data mining}, such hypotheses are represented by \emph{patterns} \cite{Aggarwal2014} and discovered by mining algorithms. Informally, a pattern is a statement in a formal language that concisely describes the structure of a subset of the data. Unfortunately, in any realistic database the \emph{interesting} and/or \emph{relevant} patterns tend to get lost among a humongous number of patterns.

The solutions that have been proposed to address this so-called \emph{pattern explosion}, caused by enumerating \emph{all} patterns satisfying given constraints, can be roughly clustered into four categories: 1) \emph{condensed representations} \cite{Calders2006}, 2) \emph{pattern set mining} \cite{Bringmann2010}, 3) \emph{pattern sampling} \cite{Boley2009}, 4) and---most recently---\emph{interactive pattern mining} \cite{VanLeeuwen2014b}. As expected, each of these categories has its own strengths and weaknesses and there is no ultimate solution as of yet.

That is, condensed representations, e.g., closed itemsets, can be lossless but usually still yield large result sets; pattern set mining and pattern sampling can provide succinct pattern sets but do not take the analyst into account; and existing interactive approaches take the user into account but do not adequately address the pattern explosion. Consequently, the analyst has to invest substantial effort in identifying those patterns that are relevant to her specific interests and goals, which often requires extensive data mining expertise.

\ParHead{Aims and contributions} Our overarching aim is to enable analysts---such as the one described in the medical scenario above---to discover small sets of patterns from data that they consider interesting. This translates to the following three specific requirements. First, we require our approach to yield \emph{concise and diverse result sets}, effectively avoiding the pattern explosion. Second, our method should \emph{take the user's interests into account} and ensure that the results are relevant. Third, it should achieve this \emph{with limited effort on behalf of the user}.

To satisfy these requirements, we propose an approach that combines pattern sampling with interactive data mining techniques. In particular, we introduce the \AlgName{} algorithm, for \textbf{Le}arn \textbf{t}o \textbf{S}ample 
\textbf{I}nteresting \textbf{P}atterns,
which follows the \emph{Mine, Interact, Learn, Repeat} framework \cite{Dzyuba2014}. It samples a small set of patterns, receives feedback from the user, exploits the feedback to learn new parameters for the sampling distribution, and repeats these steps. As a result, the user may utilize a compact diverse set of interesting patterns at any moment, blurring the boundaries between learning and discovery modes.

We satisfy the first requirement by using a sampling technique that samples high quality patterns with high probability. While sampling does not guarantee diversity \emph{per se}, we demonstrate that it gives concise yet diverse results in practice. Moreover, sampling has the advantage that it is \emph{anytime}, i.e., the result set can grow by user's request. \AlgName{}'s sampling component is based on recent advances in sampling in SAT \cite{Chakraborty2014} and their extension to pattern sampling \cite{Dzyuba2016}.

The second requirement is satisfied by learning what matters to the user, i.e., by interactively learning the distribution patterns are sampled from. This allows the user to steer the sampler towards subjectively interesting regions. We build upon recent work \cite{Dzyuba2014,Boley2013} that uses \emph{preference learning} to learn to rank patterns.

Although user effort can partially be quantified by the total amount of input that needs to be given during the analysis, the third requirement also concerns the time that is needed to find the first interesting results. For this it is of particular interest to study the \emph{trade-off between exploitation and exploration}. As mentioned, one of the benefits of interactive pattern sampling is that the boundaries between learning and discovery are blurred, meaning that the system keeps learning while it continuously aims to discover potentially interesting patterns.

We evaluate the performance of the proposed algorithm and compare it to the state-of-the-art in interactive pattern mining by emulating the interests of a user. The results confirm that the proposed algorithm has the capacity to learn
what matters based on little feedback from the user. More importantly, the \AlgName{} algorithm demonstrates favourable trade-offs concerning both quality--diversity and exploitation--exploration when compared to existing methods.
\section{Interactive pattern mining: Problem definition}%
\label{sec:problem}

Recall the medical analyst example.
We assume that after 
inspecting patterns,
she can judge their interestingness, 
e.g., by comparing two patterns.
Then the primary task of
interactive pattern mining
consists in learning
a formal model of her interests.
The second task involves
using this model to mine
novel patterns that are
subjectively interesting
to the user (according to 
the learned model).

Formally, let $\DS$ denote a dataset,
$\PattLang$ a pattern language,
$\Constraints$ a (possibly empty) 
set of constraints on patterns,
and $\succ$ the unknown subjective 
pattern preference relation of
the current user over $\PattLang$, 
i.e., $\Ptn_1 \succ \Ptn_2$
implies that the user considers
pattern $\Ptn_1$ subjectively
more interesting than pattern $\Ptn_2$:

\begin{problem}[Learning]
Given $\DS$, $\PattLang$, 
and $\Constraints$, 
dynamically collect
feedback $\Feedback$
with respect to 
patterns in $\PattLang$
and use $\Feedback$ 
to learn a (subjective)
pattern interestingness function 
$\FnDef{\Hyp}{\PattLang}{\Reals}$
such that $\Fn{\Hyp}{\Ptn_1} > \Fn{\Hyp}{\Ptn_1}%
	\Leftrightarrow \Ptn_1 \succ \Ptn_2$.
\end{problem}

The mining task should 
account for the potential diversity 
of user's interests.
For example, the analyst 
may (unwittingly) be 
interested in several
unrelated treatments
with disparate latent factors.
An algorithm should be able 
to identify and mine
patterns that are
representative of these 
diverse hypotheses.

\begin{problem}[Mining]
Given $\DS$, $\PattLang$, 
$\Constraints$, and $\Hyp$, 
mine a set of patterns 
$\PattSet_{\Hyp}$ that
maximizes a combination of
interestingness $\Hyp$ 
and diversity of patterns.

\smallskip \noindent%
The interestingness of $\PattSet$ 
can be quantified by
the average quality of 
its members, i.e.,
$\left.\sum_{\Ptn \in \PattSet}%
	{\Fn{\Hyp}{\Ptn}} \mid/ %
\Abs{\PattSet}\right.$.
Diversity measures
quantify how different
patterns in a set
are from each other. 
\emph{Joint entropy} is
a common diversity measure
\cite{VanLeeuwen2013}%
\LongShort{
	(see Section~\ref{sec:prelim}
	for the definition).%
}{.}
\end{problem}
\section{Related work}

In this paper, we focus on 
two classes of related work 
aimed at alleviating 
the pattern explosion, namely
1) pattern sampling and
2) interactive pattern mining.

\ParHead{Pattern sampling.}
First pattern samplers are based on 
Markov Chain Monte Carlo (MCMC)
random walks over the pattern lattice
\cite{Boley2009,Hasan2009,Boley2010}.
Their main advantage is that
they support ``black box''
distributions, i.e., they do not
require any prior knowledge about
the target distribution,
a property essential for
interactive exploration.
However, they often converge only slowly 
to the desired target distribution
and require the selection of 
the ``right'' proposal distributions.

Samplers that are based on
alternative approaches include
direct two-step samplers and
XOR samplers.
Two-step samplers
\LongShort{\cite{Boley2011,Boley2012}}%
	{\cite{Boley2012}}, 
while provably accurate and efficient,
only support a limited number 
of distributions and thus
cannot be easily extended
to interactive settings.
\Flexics{} \cite{Dzyuba2016} is
a recently proposed pattern sampler
based on the latest advances in
weighted constrained sampling 
in SAT \cite{Chakraborty2014}.
It supports black-box
target distributions,
provides guarantees
with respect to
sampling accuracy and
efficiency, and has been
shown to be competitive
with the state-of-the-art
methods described above.

\ParHead{Interactive pattern mining.}
Most recent approaches to 
interactive pattern mining are 
based on learning to rank patterns.
They first appeared in
Xin et al. \cite{Xin2006} and
Rueping \cite{Rueping2009} and
were independently extended by
Boley et al. \cite{Boley2013} and
Dzyuba et al. \cite{Dzyuba2014}.
The central idea behind
these algorithms is 
to alternate between 
mining and learning.
\Prime{} \cite{Bhuiyan2016a} 
focuses on advanced feature construction 
for interactive mining of structured data, 
e.g., sequences or graphs.

To the best of our knowledge, 
\Bah{} \cite{Bhuiyan2016}
is the only existing approach 
to interactive itemset sampling.
It uses binary feedback 
(``likes'' and ``dislikes'')
to update weights of 
individual items.
Itemsets are sampled 
proportional to the product of
weights of constituent items.
Thus, the model of user interests 
in \Bah{} is fairly restricted; 
moreover, it potentially suffers
from convergence issues typical for MCMC.
We empirically compare 
\AlgName{} with \Bah{} in 
Section~\ref{sec:experiments}.
\section{Preliminaries}%
\label{sec:prelim}

\ParHead{Pattern mining and sampling.}
We focus on \emph{itemset mining},
i.e., pattern mining for binary data.
Let $\Items=\Set{1 \ldots M}$ 
denote a set of items.
Then, a dataset $\DS$ is a bag of 
transactions over $\Items$,
where each transaction $\Transaction$ is 
a subset of $\Items$, i.e.,
$\Transaction \subseteq \Items$;
$\Transactions=\Set{1 \ldots N}$ is
a set of transaction indices.
The pattern language $\PattLang$ 
also consists of sets of items, 
i.e., $\PattLang = 2^{\Items}$.
An itemset $\Pattern$ occurs in 
a transaction $\Transaction$, 
iff $\Pattern \subseteq \Transaction$.
The frequency of $\Pattern$ is
the proportion of transactions
in which it occurs, i.e.,
$\Freq{\Pattern} = \Abs{%
	\Set{\Transaction \in \DS \mid%
	\Pattern \subseteq \Transaction}%
} / N$.
In labeled datasets, 
each transaction $\Transaction$
has a label from $\Set{-,+}$;
$freq^{-,+}$ are defined accordingly.

\LongOnly{%
Given an (arbitrarily ordered)
pattern set $\PattSet$ of size $k$,
its diversity can be measured 
using \emph{joint entropy} $H_J$,
which essentially quantifies
the overlap of sets of transactions,
in which the patterns in $\PattSet$ occur.
Let $\Iverson{\cdot}$ 
denote the Iverson bracket,
$b^\prime \in \Set{0,1}^k$
a binary $k$-tuple, and
$\Prob{b^\prime} = \dfrac{1}{\Abs{\DS}}\ %
	\sum\limits_{\Transaction \in \DS}{%
		\prod\limits_{i \in \Range{1}{k}}%
			{\Iverson{b_{i}^{\prime} = 1 \Leftrightarrow %
			 \PattSet_i \subseteq \Transaction}}%
	}$
the fraction of
transactions in $\DS$
covered only by patterns in $\PattSet$
that correspond to 
non-zero elements of $b^\prime$
(e.g., if $k=3$ and $b^\prime = 101$,
we only count the transactions
covered by the 1st and 
the 3rd pattern and \emph{not}
covered by the 2nd pattern).
Joint entropy $H_J$ is defined as
$\Fn{H_J}{\PattSet} = %
	-\sum\limits_{b \in \Set{0,1}^k}
		{\Prob{b} \times \log{\Prob{b}}}$.
$H_J$ is measured in bits and
bounded from above by $k$.
The higher the joint entropy,
the more diverse are
the patterns in $\PattSet$
in terms of their
occurrences in $\DS$.%
}

The choice of constraints 
and a quality measure
allows a user to express 
her analysis requirements.
The most common constraint 
is \emph{minimal frequency}
$\Freq{\Pattern} \geq \MinSup$.
In contrast to hard constraints,
quality measures are used 
to describe soft preferences 
that allow to rank patterns;
see Section~\ref{sec:experiments}
for examples.

While common mining algorithms
return the top-$k$ patterns 
w.r.t. a measure
$\FnDef{\Qual}{\PattLang}{\Reals^+}$,
pattern sampling is
a randomized procedure 
that `mines' a pattern with 
probability proportional 
to its quality, i.e.,
$\Prob[\Qual]{\Pattern\text{\small\ is sampled}}=%
	\left.\Fn{\Qual}{\Pattern} \middle/ Z_{\Qual}\right.$,
if $\Ptn \in \PattLang$ 
satisfies $\Constraints$,
and $0$ otherwise, where
$Z_{\Qual}$ is the (unknown) 
normalization constant.
This is an instance of 
weighted constrained sampling.

\ParHead{Weighted constrained sampling.}
This problem has been
extensively studied in
the context of sampling
solutions of a SAT problem%
\LongOnly{ \cite{Meel2016}}.
\Wg{} \cite{Chakraborty2014} 
is a recent algorithm for 
approximate weighted sampling in SAT.
The core idea consists of
partitioning the solution space
into a number of ``cells'' and
sampling a solution 
from a random cell.
Partitioning with desired properties
is obtained via augmenting
the SAT problem with
uniformly random 
XOR constraints (XORs).

To sample a solution,
\Wg{} dynamically estimates
the number of XORs
required to obtain a suitable cell, 
generates random XORs,
stores the solutions of
the augmented problem 
(i.e., a random cell), and 
returns a perfect 
weighted sample from the cell.
Owing to the properties 
of partitioning with 
uniformly random XORs,
\Wg{} provides theoretical 
performance guarantees
regarding quality of samples
and efficiency of
the sampling procedure.

For implementation purposes,
\Wg{} only requires
an efficient oracle that
enumerates solutions.
Moreover, it treats 
the target sampling distribution
as a black box:
it requires neither
a compact description thereof, 
nor the knowledge of the
normalization constant.
Both features are crucial in
pattern sampling settings.
\Flexics{} \cite{Dzyuba2016}, 
a recently proposed
pattern sampler based on \Wg{}, 
has been shown to be 
accurate and efficient.
\LongShort{%
	See Appendix~\ref{sec:appendix} 
	for a more detailed description
	of these algorithms.
}{%
	Due to the page limit,
	we postpone further details to
	the extended version of
	this paper \cite[Appendix A]{Dzyuba2017}.
}

\ParHead{Preference learning.}
The problem of learning
ranking functions is known 
as \emph{object ranking}%
\LongOnly{ \cite{Kamishima2011}}.
A common solving technique
involves minimizing
pairwise loss, e.g.,
the number of discordant pairs.
For example, user feedback
$\Feedback = \Set{%
p_1 \succ p_3 \succ p_2,\ 
p_4 \succ p_2}$ is seen as
$\Set{%
	\Tuple{p_1 \succ p_3},
	\Tuple{p_1 \succ p_2},
	\Tuple{p_3 \succ p_2},
	\Tuple{p_4 \succ p_2}%
}$. Given feature representations
of objects $\vec{p_i}$,
object ranking is
equivalent to positive-only
classification of difference vectors, 
i.e., a ranked pair example 
$p_i~\succ~p_j$ 
corresponds to
a classification example
$\Tuple{\vec{p_i} - \vec{p_j}, +}$.
All pairs comprise a training dataset
for a scoring classifier.
Then, the predicted ranking of
any set of objects can be
obtained by sorting these objects
by classifier score descending.
For example, this formulation
is adopted by \Algo{SvmRank}
\cite{Joachims2002}.
\section{Algorithm}%
\label{sec:algorithm}

\begin{algorithm*}[t]
\begin{algorithmic}[1]
\Require Dataset $\DS$, 
	minimal frequency threshold $\MinSup$
\Ensure Query size $k$,
	query retention $\QO$,
	range $\A$,
	cell sampling strategy $\WgCellSmp$
\Statex SCD: regularisation parameter $\lambda$,
	iterations $T$;
	\Flexics{}: error tolerance $\kappa$
\Statex \Comment{\emph{Initialization}}
\State Ranking function $\Hyp_0 =$ \Call{Logistic}{$\vec{0}$, $\A$}
	\Comment Zero weights lead to uniform sampling
\State Feedback $\Feedback \leftarrow \emptyset$,
	$\Q^{*}_0 \leftarrow \emptyset$
\Statex \Comment{%
	\emph{Mine}, \emph{Interact}, \emph{Learn}, \emph{Repeat} loop}
\For{$t = 1,2,\ldots$}
	\State $R =$ \Call{TakeFirst}{$\Q^{*}_{t-1}$, $l$}
		\Comment Retain top patterns from the previous iteration
		\label{line:retain}
	\State Query $\Q_t \leftarrow R\ \cup$ \Call{SamplePatterns}{$\Hyp_{t-1}$}
		$\times \left(k-\Abs{R}\right)$ times
		\label{line:sample}
	\State $\Q^{*}_{t} =$ \Call{Order}{$\Q_t$},
		$\Feedback \leftarrow \Feedback \cup \Q^{*}_{t}$
		\Comment{Ask user to order patterns in $\Q_t$}
	\State $\Hyp_t \leftarrow $ \Call{Logistic}{%
			\Algo{LearnWeights}%
			$\left(\Feedback; \lambda, T\right)$, $\A$%
		}
\EndFor
\Function{SamplePatterns}{Sampling weight function $\FnDef{\Weight}{\PattLang}{\Range{A}{1}}$}%
	\State $C =$ \Call{\Flexics RandomCell}
		{$\DS$, $\Freq{\cdot} \geq \MinSup$, $\Weight$; $\kappa$} 
	\If{$\WgCellSmp =$ \Call{Top}{$m$}} \Return
		$m$ highest-weighted patterns \label{line:cell-top}
	\ElsIf{$\WgCellSmp =$ \Algo{Random}} \Return
		\Call{PerfectSample}{$C$, $\Weight$} \label{line:cell-wg}
	\EndIf
\EndFunction
\end{algorithmic} \caption{\AlgName{}}
\label{algo:intsamp}
\end{algorithm*}

Key questions concerning 
instantiations of the 
\emph{Mine, interact, learn, repeat}
framework include 
1) the feedback format,
2) learning quality measures from feedback,
3) mining with learned measures, 
and crucially, 
4) selecting the patterns 
to show to the user.
As pattern sampling has been 
shown to be effective in 
mining and learning,
we present \AlgName{},
a sampling-based instantiation
of the framework which 
employs \Flexics{}.
The sequel describes
the mining and learning
components of \AlgName{}.
Algorithm~\ref{algo:intsamp} 
shows its pseudocode.

\ParHead{Mining patterns by sampling.}
Recall that the main goal is
to discover patterns that are
subjectively interesting
to a particular user.
We use parameterised
logistic functions to measure
the interestingness/quality
of a given pattern $\Ptn$:
\begin{equation*}
\Fn{\QualLog}{\Ptn;\Coeff,\A} =
	\A + \frac{1 - A}{1 + \Exp{-\Vec{\Coeff} \cdot \Vec{\Ptn}}}
\end{equation*}
where $\Vec{\Ptn}$ is
the vector of pattern 
features for $\Ptn$,
$\Vec{\Coeff}$ are feature weights,
and $\A$ is a parameter
that controls the range of
the interestingness measure, 
i.e. $\QualLog \in \ORange{A}{1}$.
Examples of pattern features include
$\Fn{Length}{\Ptn} = \Abs{\Ptn}/\Abs{\Items}$, 
$\Fn{Frequency}{\Ptn} = \Freq{\Ptn}/\Abs{\DS}$,
$\Fn{Items}{\Item,\Ptn} = \Iverson{\Item \in \Ptn}$; and
$\Fn{Transactions}{\Transaction, \Ptn} = \Iverson{\Ptn \subseteq \Transaction}$,
where $\Iverson{\cdot}$ denotes
the Iverson bracket.
Weights reflect feature contributions to
pattern interestingness, e.g., 
a user might be interested in 
combinations of particular items or 
disinterested in particular transactions.
The set of features would typically 
be chosen by the mining system designer 
rather than by the user herself.
We empirically evaluate
several feature combinations
in Section~\ref{sec:experiments}.

Specifying feature weights 
manually is tedious and opaque,
if at all possible.
Below we present an algorithm
that learns the weights based on
easy-to-provide feedback
with respect to patterns.
This motivates our choice
of logistic functions: 
they enable efficient learning.
Furthermore, their bounded range  
$\Range{\A}{1}$ yields distributions 
that allow efficient sampling
directly proportional to $\QualLog$
with \Flexics{}.
Parameter $\A$ essentially
controls the \emph{tilt}
of the distribution
\cite{Dzyuba2016}.

\ParHead{User interaction \& learning from feedback.}
Following previous research
\cite{Dzyuba2014},
we use \emph{ordered feedback},
where a user is asked
to provide a total order
over a (small) number of
patterns according to
their subjective interestingness;
see Figure~\ref{fig:example}
for an example.
We assume that there exists 
an unknown, user-specific 
target ranking $\Ranking^*$, 
i.e., a total order over $\PattLang$. 
The inductive bias is that
there exists $\vec{\Coeff}^{*}$
such that $p \succ q \Rightarrow
\Fn{\QualLog}{p, \vec{\Coeff}^{*}} > \Fn{\QualLog}{q, \vec{\Coeff}^{*}}$.
We apply the reduction 
of object ranking
to binary classification of
difference vectors
(see Section~\ref{sec:prelim}).
Following Boley et al.
\cite{Boley2013}, we use 
Stochastic Coordinate Descent (SCD) 
\cite{ShalevShwartz2011}
for minimizing L1-regularized
logistic loss.
However, unlike Boley et al.,
we directly use 
the learned functions 
for sampling.

SCD is an anytime 
convex optimization algorithm, 
which makes it suitable for
the interactive setting.
Its runtime scales linearly with 
the number of training pairs and
the dimensionality of feature vectors.
It has two parameters:
1) the number of weight updates 
(per iteration of \AlgName{}) $T$ and
2) the regularization parameter $\lambda$.
However, direct learning of $\QualLog$ 
is infeasible, as it results in
a non-convex loss function.
We therefore use SCD to optimize
the standard logistic loss, 
which is convex, and
use the learned weights 
$\Vec{\Coeff}$ in $\QualLog$.

\ParHead{Selecting patterns to show to the user.}
An interactive system seeks 
to ensure faster learning 
of accurate models
by targeted selection of patterns 
to show to the user; this is
known as \emph{active learning}
or \emph{query selection}.
Randomized methods have been 
successfully applied 
to this task \cite{Dzyuba2014}.
Furthermore, in large pattern spaces
the probability that
two redundant patterns are 
sampled in one (small) batch
is typically low.
Therefore, a sampler, which
produces independent samples, 
typically ensures diversity 
within batches and thus 
sufficient \emph{exploration}.
We directly show $\QS$ patterns 
sampled by \Flexics{}
proportional to $\QualLog$ 
to the user, for which 
she has to provide a total order
as feedback.

We propose two modifications
to \Flexics{}, which aim at
emphasising \emph{exploitation}, 
i.e., biasing sampling towards 
higher-quality patterns.
First, we employ alternative
cell sampling strategies.
Normally \Flexics{} draws 
a perfect weighted random sample,
once it obtains a suitable cell.
We denote this strategy as
$\WgCellSmp=$ \Algo{Random}.
We propose an alternative strategy 
$\WgCellSmp=$ \Algo{Top($m$)}, which 
picks the $m$~highest-quality patterns
from a cell (Line~\ref{line:cell-top}
in Algorithm~\ref{algo:intsamp}).
We hypothesize that, owing to
the properties of random XOR constraints,
patterns in a cell as well as
in consecutive cells are expected
to be sufficiently diverse and thus
the modified cell sampling 
does not disrupt exploration.

Rigorous analysis of 
(unweighted) uniform sampling 
by Chakraborty et al. shows 
that re-using samples from 
a cell still ensures broad 
coverage of the solution space,
i.e., diversity of samples
\cite{Chakraborty2015}. 
Although as a downside, 
consecutive samples are not i.i.d.,
the effects are bounded in theory
and inconsequential in practice.
We use these results 
to take license to modify
the theoretically motivated
cell sampling procedure.
Although we do not present
a similar theoretical analysis
of our modifications,
we evaluate them empirically.

Second, we propose to retain
the top $\QO$ patterns from
the previous query and only
sample $\QS-\QO$ new patterns
(Lines~\ref{line:retain}--\ref{line:sample}).
This should help users
to relate the queries 
to each other and possibly
exploit the structure
in the pattern space.
\section{Experiments}%
\label{sec:experiments}

The experimental evaluation
focuses on 1) the accuracy of
the learned user models and
2) the effectiveness of 
learning and sampling.
Evaluating interactive algorithms
is challenging, for domain experts
are scarce and it is hard
to gather enough experimental data
to draw reliable conclusions.
In order to perform
extensive evaluation,
we emulate users using
(hidden) interest models,
which the algorithm is
supposed to learn from
ordered feedback only.

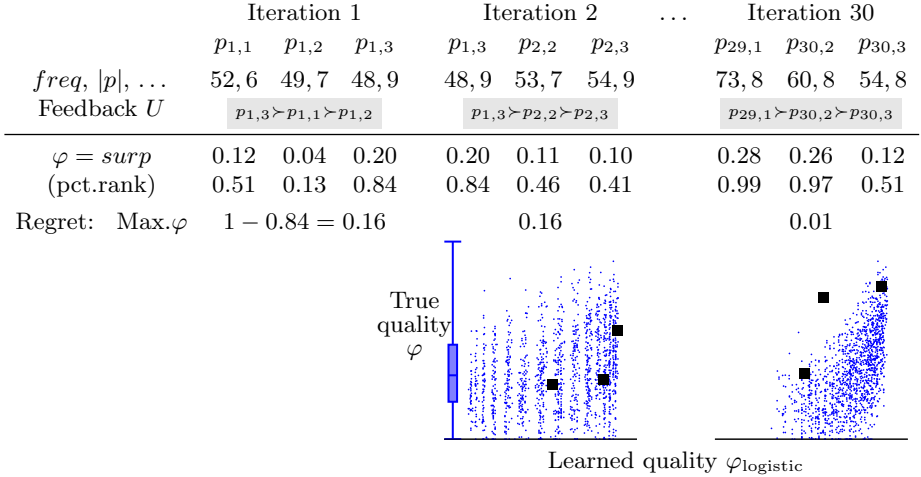
\begin{figure}[t]
\newcommand{\Feat}[2]{$#1,#2$}
\newcommand{\FeedbackW}[3]{\colorbox{gray!20}{$\scriptstyle{\Ptn_{#1} \succ \Ptn_{#2} \succ \Ptn_{#3}}$}}
\newcommand{\Scatter}[1]{\input{quallog+vs+qual-surp+lymph+48-scatter-#1}}

\setlength{\tabcolsep}{4.2pt}
\centering \begin{tabular}{ccccccccccccc}
               &              & \ColC3{Iteration 1}                        & & \ColC3{Iteration 2}                        & \ldots & \ColC3{Iteration 30}                          \\
\ColC2{}                      & $\Ptn_{1,1}$ & $\Ptn_{1,2}$ & $\Ptn_{1,3}$ & & $\Ptn_{1,3}$ & $\Ptn_{2,2}$ & $\Ptn_{2,3}$ &        & $\Ptn_{29,1}$ & $\Ptn_{30,2}$ & $\Ptn_{30,3}$ \\[3pt]               
\ColC2{{\small $\FreqW$, 
   $\Abs{\Ptn}$, \ldots}}     & \Feat{52}{6} & \Feat{49}{7} & \Feat{48}{9} & & \Feat{48}{9} & \Feat{53}{7} & \Feat{54}{9} &        & \Feat{73}{8} & \Feat{60}{8} & \Feat{54}{8}    \\[0pt]
\ColC2{Feedback $\Feedback$}  & \ColC3{\FeedbackW{1,3}{1,1}{1,2}}          & & \ColC3{\FeedbackW{1,3}{2,2}{2,3}}          &        & \ColC3{\FeedbackW{29,1}{30,2}{30,3}}          \\
\midrule
\ColC2{$\Qual=\SurpW$}        & $0.12$       & $0.04$      & $0.20$        & & $0.20$      & $0.11$      & $0.10$         &        & $0.28$      & $0.26$      & $0.12$            \\
\ColC2{(pct.rank)}            & $0.51$       & $0.13$      & $0.84$        & & $0.84$      & $0.46$      & $0.41$         &        & $0.99$      & $0.97$      & $0.51$            \\[3pt]
Regret:        & Max.$\Qual$  & \ColC3{$1-0.84=0.16$}                      & & \ColC3{$0.16$}                             &        & \ColC3{$0.01$}                                \\
               &              & \ColC3{}                                   & & \ColC3{\input{quallog+vs+qual-surp+lymph+48-scatter-0}}                        &        & \ColC3{\input{quallog+vs+qual-surp+lymph+48-scatter-28}}                          \\
\ColC6{}                                                                     & \ColC7{Learned quality $\QualLog$}                                                                  \\
\end{tabular} \caption{We emulate 
user feedback $\Feedback$ 
using a hidden quality measure $\Qual$ 
(here $\SurpW$; the boxplot shows 
the distribution of $\Qual$
in the given dataset).
The rows above the bar show
the properties of
the sampled patterns
that would be inspected by a user,
e.g., \emph{frequency} or \emph{length},
and the emulated feedback.
The scatter plots show
the relation between $\Qual$ 
and the learned model of 
user interests $\QualLog$
after 1 and 29 iterations of
feedback and learning.
The performance of
the learned model
improves considerably
as evidenced by
higher values of $\Qual$
of the sampled patterns
(squares) and lower regret.}
\label{fig:example}
\end{figure}

We follow a protocol
also used in previous work
\cite{Dzyuba2014}:
we assume that $\Ranking^*$
is derived from 
a quality measure $\Qual$, i.e., 
$p \succ q \Leftrightarrow \Fn{\Qual}{p} > \Fn{\Qual}{q}$.
Thus, the task is to learn 
to sample frequent patterns 
proportional to $\Qual$ from
(short) sample rankings.
As $\Qual$, we use 
frequency $\FreqW$, 
surprisingness $\SurpW$, and
discriminativity in labeled data
as measured by $\ChiSq$, where
$\Fn{\SurpW}{\Ptn} = \max \{\Freq{\Ptn} - \prod\limits_{\Item \in \Ptn}{\Freq{\Set{i}}}, 0\}$ and
\begin{align*}
\Fn{\ChiSq}{\Ptn} = & \sum\limits_{c \in \Set{-,+}}{%
	\frac{(\Freq{\Ptn} (\Freq[c]{\Ptn} - \Abs{\DS^{c}}))^2}{\Freq{\Ptn} \Abs{\DS^{c}}} +%
	\frac{(\Freq{\Ptn} (\Freq[c]{\Ptn} - \Abs{\DS^{c}}))^2}{(\Abs{\DS} - \Freq{\Ptn}) \Abs{\DS^{c}}}%
} \\
\end{align*}

\begin{wraptable}{r}{0.5\textwidth}
\vspace{-8pt}
\caption{Dataset properties.}

\smallskip
\setlength{\tabcolsep}{3pt}
\centering%
\begin{tabular}{lcccc}\toprule
         & \Rows2{$\Abs{\Items}$} & \Rows2{$\Abs{\DS}$} & \Rows2{$\MinSup$} & {\footnotesize Frequent} \\
\ColC3{} & & {\footnotesize patterns} \\
\midrule
\texttt{anneal}     &  $93$ &  $812$ & $660$ & $149\,331$ \\
\texttt{australian} & $125$ &  $653$ & $300$ & $141\,551$ \\
\texttt{german}     & $112$ & $1000$ & $300$ & $161\,858$ \\ 
\texttt{heart}      &  $95$ &  $296$ & $115$ & $153\,214$ \\
\texttt{hepatitis}  &  $68$ &  $137$ &  $48$ & $148\,289$ \\
\texttt{lymph}      &  $68$ &  $148$ &  $48$ & $146\,969$ \\
\texttt{primary}    &  $31$ &  $336$ &  $16$ & $162\,296$ \\
\texttt{soybean}    &  $50$ &  $630$ &  $28$ & $143\,519$ \\
\texttt{vote}       &  $48$ &  $435$ &  $25$ & $142\,095$ \\
\texttt{zoo}        &  $36$ &  $101$ &  $10$ & $151\,806$ \\
\bottomrule
\end{tabular}
\label{table:datasets}
\vspace{-25pt}
\end{wraptable} 
We investigate the performance of 
the algorithm on ten datasets\footnote{Source: 
\url{https://dtai.cs.kuleuven.be/CP4IM/datasets/}}.
For each dataset, we set
the minimal support threshold
such that there are approximately
$140\,000$ frequent patterns.
Table~\ref{table:datasets}
shows dataset statistics.
Each experiment involves
30 iterations (queries).
We use the default values 
suggested by the authors of
SCD and \Flexics{}
for the auxiliary parameters
of \AlgName{}: $\lambda=0.001$, 
$T=1000$, and $\kappa=0.9$.

We evaluate performance
using \emph{cumulative regret},
which is the difference between
the ideal value of
a certain measure $M$ 
and its observed value, 
summed over iterations.
We use the maximal and
average quality 
$\Qual$ in a query and 
joint entropy \ShortOnly{$H_J$ }%
as performance measures.
To allow comparison
across datasets and measures,
we use percentile ranks by $\Qual$
as a non-parametric measure of
ranking performance.
We also divide joint entropy
by~$k$: thus, the ideal value 
of each measure is $1$
(e.g., the highest possible
$\Qual$ over all frequent patterns
has the percentile rank of $1$),
and the regret is defined as
$\sum{1 - \Fn{M}{\Q^{*}_{i}}}$,
where $M \in \Set{\Qual_{avg}, \Qual_{max}, H_J}$.
We repeat each experiment
ten times with different 
random seeds and report 
average regret.

\ParHead{A characteristic experiment in detail.}
Figure~\ref{fig:example} illustrates
the workings of \AlgName{} and
the experimental setup.
It uses the \texttt{lymph} dataset,
the target quality measure $\Qual=\SurpW$,
\LongShort{$Items$ as features, and 
the following parameter settings:}%
	{features $=Items$,}
$\QS=3$, $\A=0.1$, $\QO=1$, 
$\WgCellSmp=$ \Algo{Random}.

\AlgName{} starts by sampling 
patterns uniformly.
A human user would inspect 
the patterns (items not shown)
and their properties, e.g., 
frequency or length, or
visualizations thereof, and
rank the patterns by their 
subjective interestingness;
in these experiments, we order them
according to their values of $\Qual$.
The algorithm uses the feedback 
to update $\QualLog$.
At the next iteration,
the patterns are sampled
from an updated distribution.
As $\QO=1$, the top-ranked pattern 
from the previous iteration
($\Ptn_{1,3}$) is retained.
After a number of iterations,
the accuracy of the approximation
increases considerably, while
the regret decreases.
On average, one iteration
takes $0.5$s on
a desktop computer.

\ParHead{Evaluating components of \AlgName{}.}
We investigate the effects of 
the choice of features and 
parameter values on
the performance of \AlgName{},
in particular query size~$\QS$,
query retention~$\QO$, range~$\A$, and 
cell sampling strategy $\WgCellSmp$.
We use the following
feature combinations
($\Vert$ denotes concatenation):
\emph{Items} (I);
\emph{Items}$\Vert$\emph{Length}$\Vert$\emph{Frequency} (ILF); and
\emph{Items}$\Vert$\emph{Length}$\Vert$\emph{Frequency}$\Vert$\emph{Transactions} (ILFT).
Values for other parameters 
and aggregated results are
shown in Table~\ref{table:params-results}.

Increasing the query size
decreases the maximal quality regret 
more than twofold, which indicates that
the proposed learning technique
is able to identify
the properties of target measures
from ordered lists of patterns.
However, as larger queries 
also increase the user effort,
further we use a more reasonable
query size of $\QS=5$.
Similarly, additional features
provide valuable information
to the learner.
Changing the range $\A$ 
does not affect the performance.

The choice of values for
query retention $\QO$ and
the cell sampling strategy
allows influencing
the exploration-exploitation trade-off.
Interestingly, retaining
one highest-ranked pattern results 
in the lowest regret with respect to
the \emph{maximal} quality.
Fully random queries ($\QO=0$)
do not enable sufficient exploitation,
whereas higher retention ($\QO\geq2$)---while 
ensuring higher \emph{average} quality%
---prevents exploration necessary 
for learning accurate weights.

The cell sampling strategy is
the only parameter that clearly 
affects joint entropy, with 
purely random cell sampling
yielding the lowest regret.
However, it is also results in 
the highest quality regrets, which
negates the gains in diversity.
Taking the best pattern
according to $\QualLog$ ensures 
the lowest quality regrets and
joint entropy equivalent
to other strategies.
Based on these findings,
we use the following parameters 
in the remaining experiments:
$\QS=5$, features $=$ ILFT,
$\A=0.5$, $\QO=1$, $\WgCellSmp=$ \Algo{Top(1)}.

The largest proportion of
\AlgName{}'s runtime costs is 
associated with sampling
(costs of weight learning are 
low due to a relatively
low number of examples).
The most important factor is
the number of items $\Abs{\Items}$:
the average runtime per iteration
ranges from 0.8s for \texttt{lymph}
to 5.8s for \texttt{australian},
which is suitable for
online data exploration.
\LongOnly{See the \Flexics{} paper
\cite{Dzyuba2016}
for more information 
about the scalability of
the sampling component.}

\begin{table}[t]
\caption{Effect of \AlgName{}'s
parameters on regret 
w.r.t. three performance measures.
Results are aggregated over 
datasets, quality measures, 
and other parameters.}
\setlength{\tabcolsep}{7.5pt}
\centering \begin{tabular}{llccc}\toprule
                                   &                        & Regret: avg.$\Qual$         & Regret: max.$\Qual$     & Regret: $H_J$ \\
\midrule
\Rows2{Query size $\QS$}           & 5                      &     \MeanSD{\HPh6.35}{1.04} &     \MeanSD{1.13}{0.52} & \MeanSDBest{13.28}{0.89} \\
                                   & \textbf{10}            & \MeanSDBest{\HPh5.91}{0.59} & \MeanSDBest{0.47}{0.18} &     \MeanSD{17.44}{0.45} \\
\midrule
\ColC5{\scriptsize All results below are for query size of $\QS=5$} \\
\midrule
\Rows3{Features}                   & I                      &     \MeanSD{\HPh8.17}{0.96} &     \MeanSD{1.35}{0.56} &     \MeanSD{13.64}{0.90} \\
                                   & ILF                    &     \MeanSD{\HPh6.30}{1.36} &     \MeanSD{1.16}{0.59} &     \MeanSD{13.15}{0.96} \\
                                   & \textbf{ILFT}          & \MeanSDBest{\HPh4.60}{0.78} & \MeanSDBest{0.87}{0.40} & \MeanSDBest{13.06}{0.81} \\
\midrule
\Rows2{Range $\A$}                 & 0.5                    &     \MeanSD{\HPh6.43}{1.06} &     \MeanSD{1.15}{0.52} & \MeanSDBest{13.20}{0.86} \\
                                   & \textbf{0.1}           & \MeanSDBest{\HPh6.26}{1.01} & \MeanSDBest{1.11}{0.51} &     \MeanSD{13.36}{0.91} \\
\midrule
\Rows4{Query retention $\QO$}      & 0                      &     \MeanSD{\HPh8.19}{1.21} &     \MeanSD{2.53}{0.72} &     \MeanSD{13.38}{0.69} \\
                                   & \textbf{1}             &     \MeanSD{\HPh6.78}{0.99} & \MeanSDBest{0.53}{0.34} & \MeanSDBest{13.06}{0.72} \\
                                   & 2                      &     \MeanSD{\HPh5.61}{0.94} &     \MeanSD{0.61}{0.42} &     \MeanSD{13.56}{1.05} \\
                                   & 3                      & \MeanSDBest{\HPh4.80}{1.00} &     \MeanSD{0.80}{0.57} &     \MeanSD{13.33}{1.22} \\
\midrule
\Rows4{Cell sampling $\WgCellSmp$} & \Algo{Random}          &        \MeanSD{10.60}{0.71} &     \MeanSD{1.89}{0.64} & \MeanSDBest{12.15}{0.59} \\
                                   & \Algo{\textbf{Top(1)}} & \MeanSDBest{\HPh5.14}{1.13} & \MeanSDBest{0.81}{0.45} &     \MeanSD{13.70}{1.00} \\
                                   & \Algo{Top(2)}          &     \MeanSD{\HPh5.45}{1.06} &     \MeanSD{0.87}{0.47} &     \MeanSD{13.60}{0.98} \\
                                   & \Algo{Top(3)}          &     \MeanSD{\HPh5.95}{1.20} &     \MeanSD{0.95}{0.50} &     \MeanSD{13.57}{0.96} \\
\bottomrule
\end{tabular} \label{table:params-results}
\end{table}

\ParHead{Comparing with alternatives.}
We compare \AlgName{} with 
\Aples{} \cite{Dzyuba2014}, 
another approach based on
active preference learning, 
and \Bah{} \cite{Bhuiyan2016},
an MCMC-based interactive 
sampling framework.
For the former, we use
query size $k$ and
feature representation
identical to \AlgName{},
query selector \Algo{MMR%
($\alpha=0.3$, $\lambda=0.7$)}, 
$C_{\text{\Algo{RankSVM}}}=0.005$,
and 1000 frequent patterns 
sampled uniformly at random and
sorted by $\FreqW$ 
as the \emph{source ranking}.
To compute regret, we use
the top-5 frequent patterns
according to the learned 
ranking function.

To emulate binary feedback
for \Bah{} based on $\Qual$,
we use a technique similar 
to the one used by the authors:
we designate a number of items
as ``interesting'' and
``like'' an itemset, if
more than half of its items
are ``interesting''.
To select the items,
we sort frequent patterns
by $\Qual$ descending and add items 
from the top-ranked patterns
until $15\%$ of all patterns 
are considered ``liked''.

As we were not able 
to obtain the code for \Bah{},
we implemented its sampling component
by materializing all frequent patterns 
and generating perfect samples
according to the learned
multiplicative distribution.
Note that this approach
favors \Bah{}, as it eliminates
the issues of MCMC convergence.
We request 300 samples
(the amount of training data 
roughly equivalent to 
that of \AlgName{}),
partition them into
30 groups of 10 patterns each, and
use the tail 5 patterns 
in each group for 
regret calculations.
Following the authors' recommendations,
we set the learning parameter to $b = 1.75$.
For the sampling-based methods
\AlgName{} and \Bah{},
we also report 
the diversity regret
as measured by joint entropy.

Table~\ref{table:comp-results} 
shows the results.
\LongShort{
	Note that the regret 
	of \AlgName{} is lower than in 
	Table~\ref{table:params-results},
	as the specific parameter
	combination suggested by 
	the previous experiments is used.
	Furthermore, it%
}{%
	The regret of \AlgName{}%
}
is substantially lower than 
that of either of the alternatives.
The advantage over \Bah{} is 
due to a more powerful
learning mechanism and
feature representation.
\Bah{}'s multiplicative weights
are biased towards longer itemsets and
items seen at early iterations,
which may prevent 
sufficient exploration,
as evidenced by higher
joint entropy regret.
Non-sampling method
\Aples{} performs 
the best for $\Qual=\FreqW$,
which can be represented
as a linear function
of the features and
learned by \Algo{RankSVM}
with the linear kernel.
It performs substantially
worse in other settings and
has the highest variance, which
reveals the importance of 
informed source rankings 
and the cons of 
pool-based active learning.
These results validate
the design choices made 
in \AlgName{}.

\begin{table}[t]
\caption{\AlgName{} has considerably 
	lower regrets than alternatives
	w.r.t. quality and, for samplers, 
	diversity as quantified by joint entropy.
	(For $\Qual=\SurpW$ (marked by~*),
	\Bah{} fails for 7 out of 10 datasets
	due to double overflow of
	multiplicative weights.)}
\setlength{\tabcolsep}{3.5pt}
\centering \begin{tabular}{lcccrccc}\toprule
           & \ColC3{Regret: avg.$\Qual$}                                                       & & \ColC3{Regret: joint entropy $H_J$} \\[2pt]
           &              $\FreqW$ &              $\ChiSq$ &                          $\SurpW$ & &                   $\FreqW$ &               $\ChiSq$ &                           $\SurpW$ \\
\midrule
\AlgName{} &     \MeanSD{2.4}{0.5} & \MeanSDBest{2.4}{0.1} & \MeanSDBest{4.5}{1.4}\hphantom{*} & &     \MeanSDBest{11.7}{0.6} & \MeanSDBest{11.7}{0.5} & \MeanSDBest{15.9}{1.1}\hphantom{*} \\[3pt]
\Bah{}     &    \MeanSD{15.5}{1.8} &    \MeanSD{12.8}{2.3} &    \MeanSD{15.5}{1.8}*            & &         \MeanSD{15.7}{1.9} &     \MeanSD{15.4}{1.9} &     \MeanSD{19.8}{2.1}*            \\
\Aples{}   & \MeanSDBest{0.0}{0.0} &     \MeanSD{4.5}{3.8} &     \MeanSD{5.3}{3.9}\hphantom{*} & &                         -- &                     -- &                     --\hphantom{*} \\
\bottomrule
\end{tabular} \label{table:comp-results}
\end{table}
\section{Conclusion}

We presented \AlgName{}, 
a sampling-based instantiation of the 
\emph{Mine, interact, learn, repeat} 
interactive pattern mining framework. 
The user is asked to rank
small sets of patterns according to 
their (subjective) interestingness.
The learning component 
uses this feedback to build
a model of user interests
via active preference learning.
The model directly defines 
the sampling distribution, which 
assigns higher probabilities 
to more interesting patterns.
The sampling component uses
the recently proposed 
\Flexics{} sampler, which we modify 
to facilitate control over
the exploration-exploitation balance
in active learning.

We empirically demonstrate 
that \AlgName{} satisfies 
the key requirements to
an interactive mining system.
We apply it to itemset mining, 
using a well-principled method 
to emulate a user.
The results demonstrate that 
\AlgName{} learns to sample 
diverse sets of interesting patterns.
Furthermore, it outperforms
two state-of-the-art
interactive methods.
This confirms that it has 
the capacity to tackle
the pattern explosion while
taking user interests into account.

Directions for future work 
include extending \AlgName{}
to other pattern languages, 
e.g., association rules,
investigating the effect 
of noisy user feedback 
on the performance, and
formal analysis, e.g.,
with multi-armed bandits
\cite{Filippi2010}.
A user study is necessary 
to evaluate the practical aspects
of the proposed approach.

\medskip

\noindent%
\textbf{Acknowledgements}:
Vladimir Dzyuba is supported
by FWO-Vlaanderen.
The authors would like
to thank the anonymous reviewers
for their helpful feedback.
 
\bibliographystyle{splncs03}

\LongOnly{%
\appendix
\section{Sampling patterns with \Flexics{}}%
\label{sec:appendix}

Here we present a bird's eye view
on the \Wg{}/\Flexics{} 
sampling procedure in order 
to provide context for
the modifications to it 
made within \AlgName{}, which are
described in Section~\ref{sec:algorithm}.
The reader interested in
further technical details 
should consult
the respective papers
\cite{Dzyuba2016,Chakraborty2014}.

\Wg{} is an algorithm for
weighted constrained sampling of
solutions to a SAT problem $\Formula$,
where each solution $\Assignment$ is
assigned a weight;
$\WFn{\Assignment} \in \LOrange{0}{1}$.
The goal is to sample 
solutions to $\Formula$ randomly,
with the probability of sampling
a solution $\Assignment$
proportional to its weight 
$\WFn{\Assignment}$.
\Wg{} employs the following
high-level sampling procedure:
1) partition the set of solutions
to $\Formula$ into a number 
of random subsets 
(referred to as \emph{cells});
2) sample a random cell, and
3) sample a random solution 
from that cell.
The key challenges are
obtaining a partitioning
with desirable properties and 
enumerating the solutions 
in a random cell efficiently.
\Wg{} addresses these
with partitionings induced 
by random XOR constraints (XORs).

\Flexics{} extends
this sampling procedure from 
SAT to pattern mining.
Variables in XORs correspond 
to individual items:
$\bigotimes_{\Item \in \Items}%
	{b_i \cdot \Iverson{\Item \in \Ptn}} = b_0$, 
where $b_{0 \mid i} \in \Set{0,1}$.
The coefficients $b_i$ 
determine the items involved
in the constraint, whereas
the \emph{parity bit} $b_0$ 
determines whether 
an even or an odd number 
of the involved items
must be set to $1$ for
a pattern $\Ptn$
to satisfy the XOR.
Together, $m$ XORs identify 
one cell belonging
to a partitioning of
the set of all patterns
into $2^m$ cells.
A required number of
XORs is estimated
once per batch, based on
theoretical considerations.
Then for each sample 
(1) the coefficients $b$
are drawn uniformly,
obtaining a random cell;
(2) \Flexics{} enumerates
and stores all patterns 
in that cell, i.e., 
the patterns that satisfy 
the original constraints $\Constraints$
\emph{and} the sampled XORs;
(3) a perfect sample 
is drawn from the cell and
returned as the overall sample.
Theoretical properties of
uniformly drawn XORs
allow proving
desirable properties of
the partitioning and
bounding the sampling error.

\begin{figure}[t]
\centering
\begingroup
  \makeatletter
  \providecommand\color[2][]{%
    \GenericError{(gnuplot) \space\space\space\@spaces}{%
      Package color not loaded in conjunction with
      terminal option `colourtext'%
    }{See the gnuplot documentation for explanation.%
    }{Either use 'blacktext' in gnuplot or load the package
      color.sty in LaTeX.}%
    \renewcommand\color[2][]{}%
  }%
  \providecommand\includegraphics[2][]{%
    \GenericError{(gnuplot) \space\space\space\@spaces}{%
      Package graphicx or graphics not loaded%
    }{See the gnuplot documentation for explanation.%
    }{The gnuplot epslatex terminal needs graphicx.sty or graphics.sty.}%
    \renewcommand\includegraphics[2][]{}%
  }%
  \providecommand\rotatebox[2]{#2}%
  \@ifundefined{ifGPcolor}{%
    \newif\ifGPcolor
    \GPcolortrue
  }{}%
  \@ifundefined{ifGPblacktext}{%
    \newif\ifGPblacktext
    \GPblacktextfalse
  }{}%
  \let\gplgaddtomacro\g@addto@macro
  \gdef\gplbacktext{}%
  \gdef\gplfronttext{}%
  \makeatother
  \ifGPblacktext
    \def\colorrgb#1{}%
    \def\colorgray#1{}%
  \else
    \ifGPcolor
      \def\colorrgb#1{\color[rgb]{#1}}%
      \def\colorgray#1{\color[gray]{#1}}%
      \expandafter\def\csname LTw\endcsname{\color{white}}%
      \expandafter\def\csname LTb\endcsname{\color{black}}%
      \expandafter\def\csname LTa\endcsname{\color{black}}%
      \expandafter\def\csname LT0\endcsname{\color[rgb]{1,0,0}}%
      \expandafter\def\csname LT1\endcsname{\color[rgb]{0,1,0}}%
      \expandafter\def\csname LT2\endcsname{\color[rgb]{0,0,1}}%
      \expandafter\def\csname LT3\endcsname{\color[rgb]{1,0,1}}%
      \expandafter\def\csname LT4\endcsname{\color[rgb]{0,1,1}}%
      \expandafter\def\csname LT5\endcsname{\color[rgb]{1,1,0}}%
      \expandafter\def\csname LT6\endcsname{\color[rgb]{0,0,0}}%
      \expandafter\def\csname LT7\endcsname{\color[rgb]{1,0.3,0}}%
      \expandafter\def\csname LT8\endcsname{\color[rgb]{0.5,0.5,0.5}}%
    \else
      \def\colorrgb#1{\color{black}}%
      \def\colorgray#1{\color[gray]{#1}}%
      \expandafter\def\csname LTw\endcsname{\color{white}}%
      \expandafter\def\csname LTb\endcsname{\color{black}}%
      \expandafter\def\csname LTa\endcsname{\color{black}}%
      \expandafter\def\csname LT0\endcsname{\color{black}}%
      \expandafter\def\csname LT1\endcsname{\color{black}}%
      \expandafter\def\csname LT2\endcsname{\color{black}}%
      \expandafter\def\csname LT3\endcsname{\color{black}}%
      \expandafter\def\csname LT4\endcsname{\color{black}}%
      \expandafter\def\csname LT5\endcsname{\color{black}}%
      \expandafter\def\csname LT6\endcsname{\color{black}}%
      \expandafter\def\csname LT7\endcsname{\color{black}}%
      \expandafter\def\csname LT8\endcsname{\color{black}}%
    \fi
  \fi
    \setlength{\unitlength}{0.0500bp}%
    \ifx\gptboxheight\undefined%
      \newlength{\gptboxheight}%
      \newlength{\gptboxwidth}%
      \newsavebox{\gptboxtext}%
    \fi%
    \setlength{\fboxrule}{0.5pt}%
    \setlength{\fboxsep}{1pt}%
\begin{picture}(4320.00,3024.00)%
    \gplgaddtomacro\gplbacktext{%
    }%
    \gplgaddtomacro\gplfronttext{%
      \csname LTb\endcsname%
      \put(4066,1014){\makebox(0,0)[l]{\strut{}$\Qual=0.1=A$}}%
      \csname LTb\endcsname%
      \put(4066,1289){\makebox(0,0)[l]{\strut{}$\Qual=0.4\hphantom{=A}$}}%
      \csname LTb\endcsname%
      \put(4066,1564){\makebox(0,0)[l]{\strut{}$\Qual=0.7\hphantom{=A}$}}%
      \csname LTb\endcsname%
      \put(4066,1840){\makebox(0,0)[l]{\strut{}$\Qual=1.0\hphantom{=A}$}}%
    }%
    \gplbacktext
    \put(0,0){\includegraphics{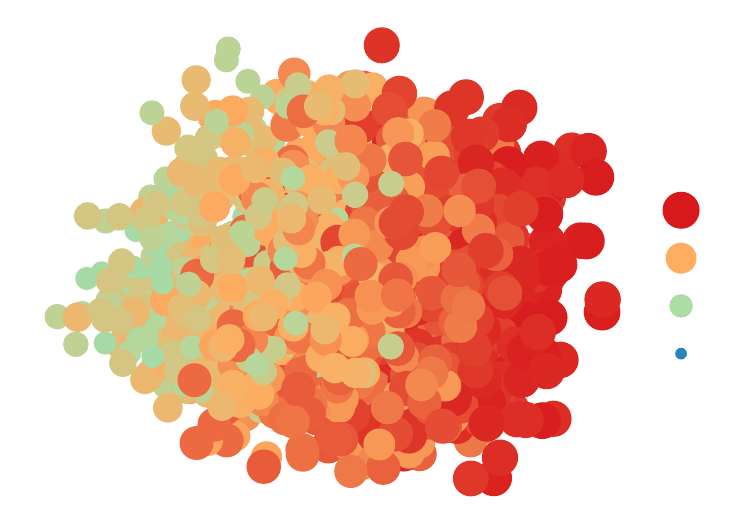}}%
    \gplfronttext
  \end{picture}%
\endgroup
 \caption{The two principal components 
obtained from the $Items$ features
of all frequent patterns,
i.e., pattern descriptions%
\protect\footnotemark.
The size and the color of a point
indicate the value of $\QualLog$ of
the corresponding pattern.
For clarity, only a $1\%$-subsample is shown.}
\label{fig:pca-all}
\end{figure}

\footnotetext{%
	The PCA coordinates 
	and $\QualLog$ are 
	strongly correlated,
	because they are computed
	using the same 
	feature representation
	for patterns ($Items$).%
}

In order to illustrate
the concepts described above,
we use the characteristic example
from Section~\ref{sec:experiments}
with $\Qual=\QualLog$ after
$30$ learning iterations.
We visualize patterns
by plotting the two 
principal components
obtained from the $Items$
feature matrix, i.e., 
pattern descriptions.
Figure~\ref{fig:pca-all}
shows all frequent patterns, 
while Figure~\ref{fig:pca-cells}
shows examples of random cells,
i.e., the output of
\Algo{FlexicsRandomCell},
from which patterns are
chosen by a cell sampling
strategy $\WgCellSmp$.

The cells are different
from each other, thus
patterns returned from
consecutive cells are
independent and diverse.
In each cell, we highlight
the pattern with
the highest quality $\QualLog$, 
which is returned by
$\WgCellSmp=$ \Algo{Top(1)}, 
along with $\ProbSymbol_{\WgCellSmp=\text{\Algo{Random}}}$, 
the probability that it is 
\emph{sampled from that cell}
if $\WgCellSmp=$ \Algo{Random}.
These probabilities do not
exceed $0.05$, which demonstrates
the motivation for alternative
cell sampling strategies.
As expected, the patterns 
returned by \Algo{Top(1)} are 
concentrated in the regions 
in the pattern space that are
characterized by high values 
of $\QualLog$. Nevertheless, 
they are different from each other,
thus the diversity across samples 
is maintained, 
regardless of the bias 
towards exploitation.

\newcommand{\CellFig}[1]{
	\input{cells-surp+lymph+48-pca-#1}%
}

\begin{figure}[p]
\setlength{\tabcolsep}{5.5pt}
\centering \begin{tabular}{cc}

	\input{cells-surp+lymph+48-pca-2}%
 & 
	\input{cells-surp+lymph+48-pca-1}%
 \\

	\input{cells-surp+lymph+48-pca-3}%
 & 
	\input{cells-surp+lymph+48-pca-6}%
 \\

	\input{cells-surp+lymph+48-pca-4}%
 & 
	\input{cells-surp+lymph+48-pca-5}%
 \\

	\input{cells-surp+lymph+48-pca-7}%
 & 
	\input{cells-surp+lymph+48-pca-8}%
 \\

	\input{cells-surp+lymph+48-pca-9}%
 & 
	\input{cells-surp+lymph+48-pca-10}%
 \\
\end{tabular}
\caption{Individual cells plotted
using the same PCA transformation
as in Figure~\ref{fig:pca-all},
i.e., the distances between patterns
correspond to the distances 
between their descriptions.}
\label{fig:pca-cells}
\end{figure}
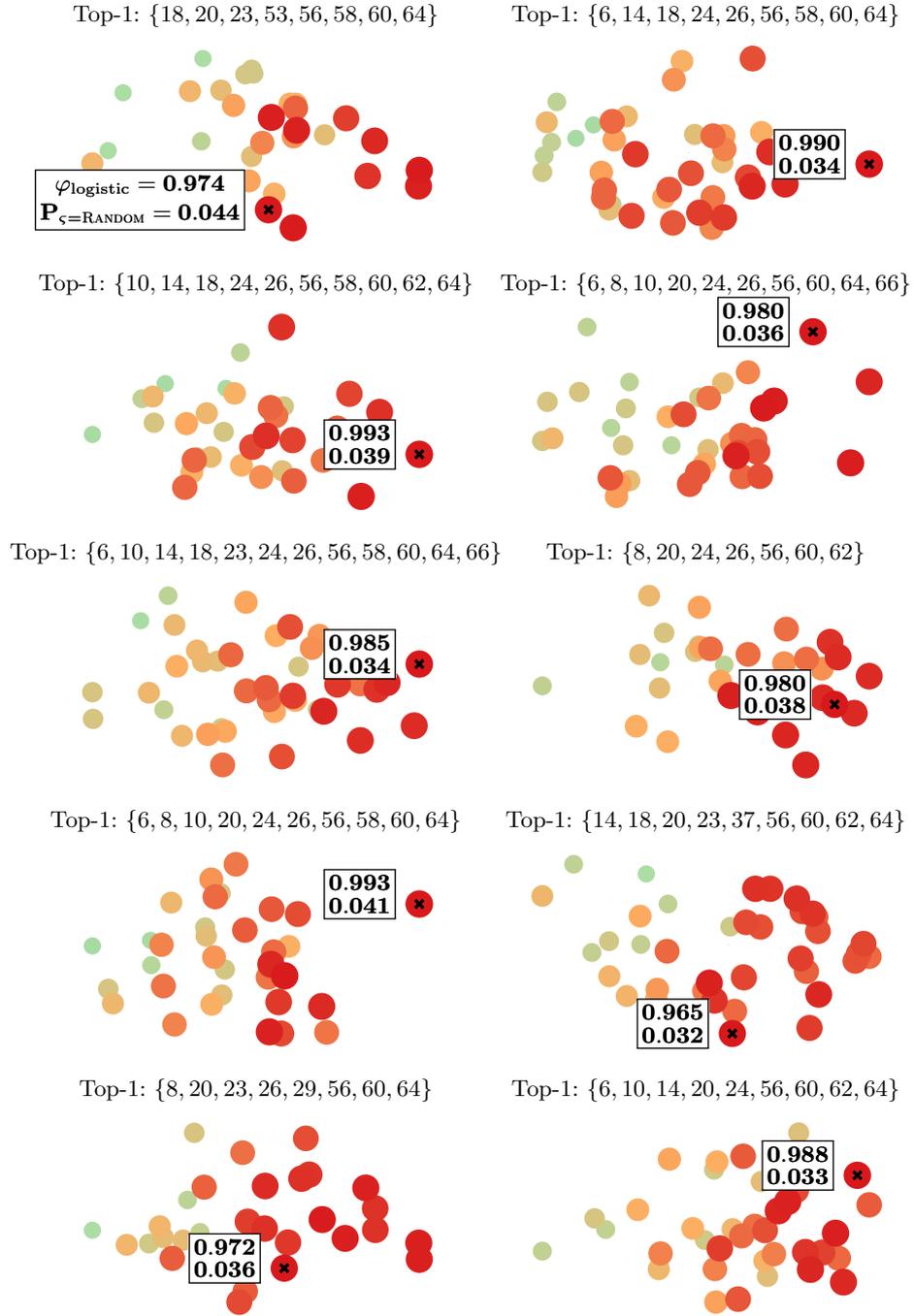
}
	
\end{document}

%% file: quallog+vs+qual-surp+lymph+48-scatter-0.tex
\begingroup
  \makeatletter
  \providecommand\color[2][]{%
    \GenericError{(gnuplot) \space\space\space\@spaces}{%
      Package color not loaded in conjunction with
      terminal option `colourtext'%
    }{See the gnuplot documentation for explanation.%
    }{Either use 'blacktext' in gnuplot or load the package
      color.sty in LaTeX.}%
    \renewcommand\color[2][]{}%
  }%
  \providecommand\includegraphics[2][]{%
    \GenericError{(gnuplot) \space\space\space\@spaces}{%
      Package graphicx or graphics not loaded%
    }{See the gnuplot documentation for explanation.%
    }{The gnuplot epslatex terminal needs graphicx.sty or graphics.sty.}%
    \renewcommand\includegraphics[2][]{}%
  }%
  \providecommand\rotatebox[2]{#2}%
  \@ifundefined{ifGPcolor}{%
    \newif\ifGPcolor
    \GPcolortrue
  }{}%
  \@ifundefined{ifGPblacktext}{%
    \newif\ifGPblacktext
    \GPblacktextfalse
  }{}%
  \let\gplgaddtomacro\g@addto@macro
  \gdef\gplbacktext{}%
  \gdef\gplfronttext{}%
  \makeatother
  \ifGPblacktext
    \def\colorrgb#1{}%
    \def\colorgray#1{}%
  \else
    \ifGPcolor
      \def\colorrgb#1{\color[rgb]{#1}}%
      \def\colorgray#1{\color[gray]{#1}}%
      \expandafter\def\csname LTw\endcsname{\color{white}}%
      \expandafter\def\csname LTb\endcsname{\color{black}}%
      \expandafter\def\csname LTa\endcsname{\color{black}}%
      \expandafter\def\csname LT0\endcsname{\color[rgb]{1,0,0}}%
      \expandafter\def\csname LT1\endcsname{\color[rgb]{0,1,0}}%
      \expandafter\def\csname LT2\endcsname{\color[rgb]{0,0,1}}%
      \expandafter\def\csname LT3\endcsname{\color[rgb]{1,0,1}}%
      \expandafter\def\csname LT4\endcsname{\color[rgb]{0,1,1}}%
      \expandafter\def\csname LT5\endcsname{\color[rgb]{1,1,0}}%
      \expandafter\def\csname LT6\endcsname{\color[rgb]{0,0,0}}%
      \expandafter\def\csname LT7\endcsname{\color[rgb]{1,0.3,0}}%
      \expandafter\def\csname LT8\endcsname{\color[rgb]{0.5,0.5,0.5}}%
    \else
      \def\colorrgb#1{\color{black}}%
      \def\colorgray#1{\color[gray]{#1}}%
      \expandafter\def\csname LTw\endcsname{\color{white}}%
      \expandafter\def\csname LTb\endcsname{\color{black}}%
      \expandafter\def\csname LTa\endcsname{\color{black}}%
      \expandafter\def\csname LT0\endcsname{\color{black}}%
      \expandafter\def\csname LT1\endcsname{\color{black}}%
      \expandafter\def\csname LT2\endcsname{\color{black}}%
      \expandafter\def\csname LT3\endcsname{\color{black}}%
      \expandafter\def\csname LT4\endcsname{\color{black}}%
      \expandafter\def\csname LT5\endcsname{\color{black}}%
      \expandafter\def\csname LT6\endcsname{\color{black}}%
      \expandafter\def\csname LT7\endcsname{\color{black}}%
      \expandafter\def\csname LT8\endcsname{\color{black}}%
    \fi
  \fi
    \setlength{\unitlength}{0.0500bp}%
    \ifx\gptboxheight\undefined%
      \newlength{\gptboxheight}%
      \newlength{\gptboxwidth}%
      \newsavebox{\gptboxtext}%
    \fi%
    \setlength{\fboxrule}{0.5pt}%
    \setlength{\fboxsep}{1pt}%
\begin{picture}(1440.00,1512.00)%
    \gplgaddtomacro\gplbacktext{%
    }%
    \gplgaddtomacro\gplfronttext{%
      \csname LTb\endcsname%
      \put(-220,837){\makebox(0,0){\strut{}\shortstack{True\\quality\\$\Qual$}}}%
    }%
    \gplbacktext
    \put(0,0){\includegraphics{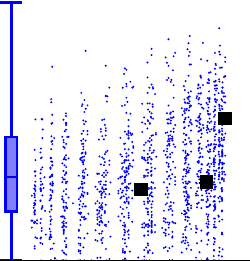}}%
    \gplfronttext
  \end{picture}%
\endgroup

%% file: quallog+vs+qual-surp+lymph+48-scatter-28.tex
\begingroup
  \makeatletter
  \providecommand\color[2][]{%
    \GenericError{(gnuplot) \space\space\space\@spaces}{%
      Package color not loaded in conjunction with
      terminal option `colourtext'%
    }{See the gnuplot documentation for explanation.%
    }{Either use 'blacktext' in gnuplot or load the package
      color.sty in LaTeX.}%
    \renewcommand\color[2][]{}%
  }%
  \providecommand\includegraphics[2][]{%
    \GenericError{(gnuplot) \space\space\space\@spaces}{%
      Package graphicx or graphics not loaded%
    }{See the gnuplot documentation for explanation.%
    }{The gnuplot epslatex terminal needs graphicx.sty or graphics.sty.}%
    \renewcommand\includegraphics[2][]{}%
  }%
  \providecommand\rotatebox[2]{#2}%
  \@ifundefined{ifGPcolor}{%
    \newif\ifGPcolor
    \GPcolortrue
  }{}%
  \@ifundefined{ifGPblacktext}{%
    \newif\ifGPblacktext
    \GPblacktextfalse
  }{}%
  \let\gplgaddtomacro\g@addto@macro
  \gdef\gplbacktext{}%
  \gdef\gplfronttext{}%
  \makeatother
  \ifGPblacktext
    \def\colorrgb#1{}%
    \def\colorgray#1{}%
  \else
    \ifGPcolor
      \def\colorrgb#1{\color[rgb]{#1}}%
      \def\colorgray#1{\color[gray]{#1}}%
      \expandafter\def\csname LTw\endcsname{\color{white}}%
      \expandafter\def\csname LTb\endcsname{\color{black}}%
      \expandafter\def\csname LTa\endcsname{\color{black}}%
      \expandafter\def\csname LT0\endcsname{\color[rgb]{1,0,0}}%
      \expandafter\def\csname LT1\endcsname{\color[rgb]{0,1,0}}%
      \expandafter\def\csname LT2\endcsname{\color[rgb]{0,0,1}}%
      \expandafter\def\csname LT3\endcsname{\color[rgb]{1,0,1}}%
      \expandafter\def\csname LT4\endcsname{\color[rgb]{0,1,1}}%
      \expandafter\def\csname LT5\endcsname{\color[rgb]{1,1,0}}%
      \expandafter\def\csname LT6\endcsname{\color[rgb]{0,0,0}}%
      \expandafter\def\csname LT7\endcsname{\color[rgb]{1,0.3,0}}%
      \expandafter\def\csname LT8\endcsname{\color[rgb]{0.5,0.5,0.5}}%
    \else
      \def\colorrgb#1{\color{black}}%
      \def\colorgray#1{\color[gray]{#1}}%
      \expandafter\def\csname LTw\endcsname{\color{white}}%
      \expandafter\def\csname LTb\endcsname{\color{black}}%
      \expandafter\def\csname LTa\endcsname{\color{black}}%
      \expandafter\def\csname LT0\endcsname{\color{black}}%
      \expandafter\def\csname LT1\endcsname{\color{black}}%
      \expandafter\def\csname LT2\endcsname{\color{black}}%
      \expandafter\def\csname LT3\endcsname{\color{black}}%
      \expandafter\def\csname LT4\endcsname{\color{black}}%
      \expandafter\def\csname LT5\endcsname{\color{black}}%
      \expandafter\def\csname LT6\endcsname{\color{black}}%
      \expandafter\def\csname LT7\endcsname{\color{black}}%
      \expandafter\def\csname LT8\endcsname{\color{black}}%
    \fi
  \fi
    \setlength{\unitlength}{0.0500bp}%
    \ifx\gptboxheight\undefined%
      \newlength{\gptboxheight}%
      \newlength{\gptboxwidth}%
      \newsavebox{\gptboxtext}%
    \fi%
    \setlength{\fboxrule}{0.5pt}%
    \setlength{\fboxsep}{1pt}%
\begin{picture}(1440.00,1512.00)%
    \gplgaddtomacro\gplbacktext{%
    }%
    \gplgaddtomacro\gplfronttext{%
    }%
    \gplbacktext
    \put(0,0){\includegraphics{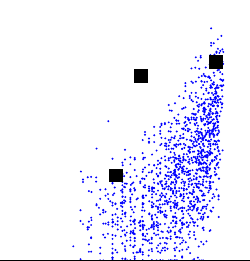}}%
    \gplfronttext
  \end{picture}%
\endgroup

%% file: cells-surp+lymph+48-pca-2.tex
\begingroup
  \makeatletter
  \providecommand\color[2][]{%
    \GenericError{(gnuplot) \space\space\space\@spaces}{%
      Package color not loaded in conjunction with
      terminal option `colourtext'%
    }{See the gnuplot documentation for explanation.%
    }{Either use 'blacktext' in gnuplot or load the package
      color.sty in LaTeX.}%
    \renewcommand\color[2][]{}%
  }%
  \providecommand\includegraphics[2][]{%
    \GenericError{(gnuplot) \space\space\space\@spaces}{%
      Package graphicx or graphics not loaded%
    }{See the gnuplot documentation for explanation.%
    }{The gnuplot epslatex terminal needs graphicx.sty or graphics.sty.}%
    \renewcommand\includegraphics[2][]{}%
  }%
  \providecommand\rotatebox[2]{#2}%
  \@ifundefined{ifGPcolor}{%
    \newif\ifGPcolor
    \GPcolortrue
  }{}%
  \@ifundefined{ifGPblacktext}{%
    \newif\ifGPblacktext
    \GPblacktextfalse
  }{}%
  \let\gplgaddtomacro\g@addto@macro
  \gdef\gplbacktext{}%
  \gdef\gplfronttext{}%
  \makeatother
  \ifGPblacktext
    \def\colorrgb#1{}%
    \def\colorgray#1{}%
  \else
    \ifGPcolor
      \def\colorrgb#1{\color[rgb]{#1}}%
      \def\colorgray#1{\color[gray]{#1}}%
      \expandafter\def\csname LTw\endcsname{\color{white}}%
      \expandafter\def\csname LTb\endcsname{\color{black}}%
      \expandafter\def\csname LTa\endcsname{\color{black}}%
      \expandafter\def\csname LT0\endcsname{\color[rgb]{1,0,0}}%
      \expandafter\def\csname LT1\endcsname{\color[rgb]{0,1,0}}%
      \expandafter\def\csname LT2\endcsname{\color[rgb]{0,0,1}}%
      \expandafter\def\csname LT3\endcsname{\color[rgb]{1,0,1}}%
      \expandafter\def\csname LT4\endcsname{\color[rgb]{0,1,1}}%
      \expandafter\def\csname LT5\endcsname{\color[rgb]{1,1,0}}%
      \expandafter\def\csname LT6\endcsname{\color[rgb]{0,0,0}}%
      \expandafter\def\csname LT7\endcsname{\color[rgb]{1,0.3,0}}%
      \expandafter\def\csname LT8\endcsname{\color[rgb]{0.5,0.5,0.5}}%
    \else
      \def\colorrgb#1{\color{black}}%
      \def\colorgray#1{\color[gray]{#1}}%
      \expandafter\def\csname LTw\endcsname{\color{white}}%
      \expandafter\def\csname LTb\endcsname{\color{black}}%
      \expandafter\def\csname LTa\endcsname{\color{black}}%
      \expandafter\def\csname LT0\endcsname{\color{black}}%
      \expandafter\def\csname LT1\endcsname{\color{black}}%
      \expandafter\def\csname LT2\endcsname{\color{black}}%
      \expandafter\def\csname LT3\endcsname{\color{black}}%
      \expandafter\def\csname LT4\endcsname{\color{black}}%
      \expandafter\def\csname LT5\endcsname{\color{black}}%
      \expandafter\def\csname LT6\endcsname{\color{black}}%
      \expandafter\def\csname LT7\endcsname{\color{black}}%
      \expandafter\def\csname LT8\endcsname{\color{black}}%
    \fi
  \fi
    \setlength{\unitlength}{0.0500bp}%
    \ifx\gptboxheight\undefined%
      \newlength{\gptboxheight}%
      \newlength{\gptboxwidth}%
      \newsavebox{\gptboxtext}%
    \fi%
    \setlength{\fboxrule}{0.5pt}%
    \setlength{\fboxsep}{1pt}%
\begin{picture}(3240.00,2016.00)%
    \gplgaddtomacro\gplbacktext{%
    }%
    \gplgaddtomacro\gplfronttext{%
      \csname LTb\endcsname%
      \put(1586,1906){\makebox(0,0){\strut{}Top-$1$: $\Set{18,20,23,53,56,58,60,64}$}}%
      \csname LTb\endcsname%
      \put(1755,1004){\makebox(0,0)[l]{\strut{}}}%
      \csname LTb\endcsname%
      \settowidth{\gptboxwidth}{\widthof{\shortstack{$\QualLog{}=\mathbf{0.974}$\\$\mathbf{\ProbSymbol_{\WgCellSmp=\text{\Algo{Random}}}}=\mathbf{0.044}$}}}
	\advance\gptboxwidth by 2\fboxsep
      \savebox{\gptboxtext}{\parbox[c][\totalheight+2\fboxsep]{\gptboxwidth}{\centering{\shortstack{$\QualLog{}=\mathbf{0.974}$\\$\mathbf{\ProbSymbol_{\WgCellSmp=\text{\Algo{Random}}}}=\mathbf{0.044}$}}}}
	\put(1504,435){\makebox[-\width][c]{\colorbox{white}{\usebox{\gptboxtext}}}}
	\settowidth{\gptboxwidth}{\usebox{\gptboxtext}}
	\advance\gptboxwidth by 2\fboxsep
	\put(1504,435){\makebox[-\width][c]{\framebox[\gptboxwidth][c]{\usebox{\gptboxtext}}}}
    }%
    \gplbacktext
    \put(0,0){\includegraphics{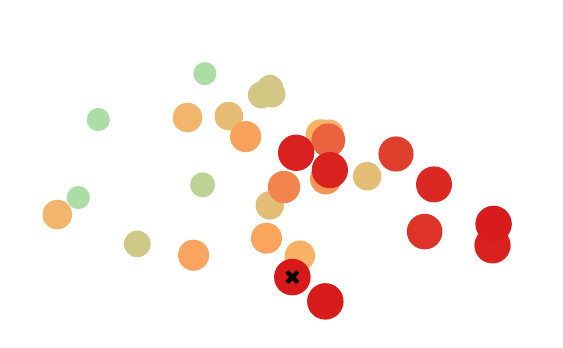}}%
    \gplfronttext
  \end{picture}%
\endgroup

%% file: cells-surp+lymph+48-pca-1.tex
\begingroup
  \makeatletter
  \providecommand\color[2][]{%
    \GenericError{(gnuplot) \space\space\space\@spaces}{%
      Package color not loaded in conjunction with
      terminal option `colourtext'%
    }{See the gnuplot documentation for explanation.%
    }{Either use 'blacktext' in gnuplot or load the package
      color.sty in LaTeX.}%
    \renewcommand\color[2][]{}%
  }%
  \providecommand\includegraphics[2][]{%
    \GenericError{(gnuplot) \space\space\space\@spaces}{%
      Package graphicx or graphics not loaded%
    }{See the gnuplot documentation for explanation.%
    }{The gnuplot epslatex terminal needs graphicx.sty or graphics.sty.}%
    \renewcommand\includegraphics[2][]{}%
  }%
  \providecommand\rotatebox[2]{#2}%
  \@ifundefined{ifGPcolor}{%
    \newif\ifGPcolor
    \GPcolortrue
  }{}%
  \@ifundefined{ifGPblacktext}{%
    \newif\ifGPblacktext
    \GPblacktextfalse
  }{}%
  \let\gplgaddtomacro\g@addto@macro
  \gdef\gplbacktext{}%
  \gdef\gplfronttext{}%
  \makeatother
  \ifGPblacktext
    \def\colorrgb#1{}%
    \def\colorgray#1{}%
  \else
    \ifGPcolor
      \def\colorrgb#1{\color[rgb]{#1}}%
      \def\colorgray#1{\color[gray]{#1}}%
      \expandafter\def\csname LTw\endcsname{\color{white}}%
      \expandafter\def\csname LTb\endcsname{\color{black}}%
      \expandafter\def\csname LTa\endcsname{\color{black}}%
      \expandafter\def\csname LT0\endcsname{\color[rgb]{1,0,0}}%
      \expandafter\def\csname LT1\endcsname{\color[rgb]{0,1,0}}%
      \expandafter\def\csname LT2\endcsname{\color[rgb]{0,0,1}}%
      \expandafter\def\csname LT3\endcsname{\color[rgb]{1,0,1}}%
      \expandafter\def\csname LT4\endcsname{\color[rgb]{0,1,1}}%
      \expandafter\def\csname LT5\endcsname{\color[rgb]{1,1,0}}%
      \expandafter\def\csname LT6\endcsname{\color[rgb]{0,0,0}}%
      \expandafter\def\csname LT7\endcsname{\color[rgb]{1,0.3,0}}%
      \expandafter\def\csname LT8\endcsname{\color[rgb]{0.5,0.5,0.5}}%
    \else
      \def\colorrgb#1{\color{black}}%
      \def\colorgray#1{\color[gray]{#1}}%
      \expandafter\def\csname LTw\endcsname{\color{white}}%
      \expandafter\def\csname LTb\endcsname{\color{black}}%
      \expandafter\def\csname LTa\endcsname{\color{black}}%
      \expandafter\def\csname LT0\endcsname{\color{black}}%
      \expandafter\def\csname LT1\endcsname{\color{black}}%
      \expandafter\def\csname LT2\endcsname{\color{black}}%
      \expandafter\def\csname LT3\endcsname{\color{black}}%
      \expandafter\def\csname LT4\endcsname{\color{black}}%
      \expandafter\def\csname LT5\endcsname{\color{black}}%
      \expandafter\def\csname LT6\endcsname{\color{black}}%
      \expandafter\def\csname LT7\endcsname{\color{black}}%
      \expandafter\def\csname LT8\endcsname{\color{black}}%
    \fi
  \fi
    \setlength{\unitlength}{0.0500bp}%
    \ifx\gptboxheight\undefined%
      \newlength{\gptboxheight}%
      \newlength{\gptboxwidth}%
      \newsavebox{\gptboxtext}%
    \fi%
    \setlength{\fboxrule}{0.5pt}%
    \setlength{\fboxsep}{1pt}%
\begin{picture}(3240.00,2016.00)%
    \gplgaddtomacro\gplbacktext{%
    }%
    \gplgaddtomacro\gplfronttext{%
      \csname LTb\endcsname%
      \put(1586,1906){\makebox(0,0){\strut{}Top-$1$: $\Set{6,14,18,24,26,56,58,60,64}$}}%
      \csname LTb\endcsname%
      \put(1592,876){\makebox(0,0)[l]{\strut{}}}%
      \csname LTb\endcsname%
      \settowidth{\gptboxwidth}{\widthof{\shortstack{$\mathbf{0.990}$\\$\mathbf{0.034}$}}}
	\advance\gptboxwidth by 2\fboxsep
      \savebox{\gptboxtext}{\parbox[c][\totalheight+2\fboxsep]{\gptboxwidth}{\centering{\shortstack{$\mathbf{0.990}$\\$\mathbf{0.034}$}}}}
	\put(2663,789){\makebox[-\width][c]{\colorbox{white}{\usebox{\gptboxtext}}}}
	\settowidth{\gptboxwidth}{\usebox{\gptboxtext}}
	\advance\gptboxwidth by 2\fboxsep
	\put(2663,789){\makebox[-\width][c]{\framebox[\gptboxwidth][c]{\usebox{\gptboxtext}}}}
    }%
    \gplbacktext
    \put(0,0){\includegraphics{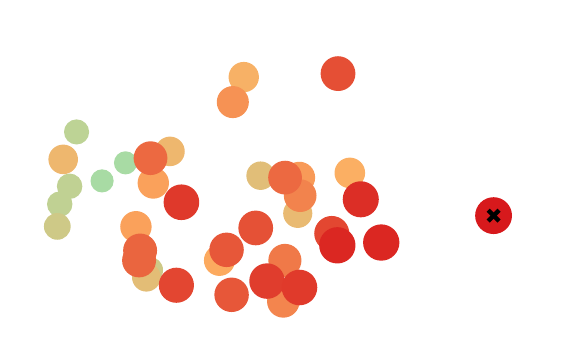}}%
    \gplfronttext
  \end{picture}%
\endgroup

%% file: cells-surp+lymph+48-pca-3.tex
\begingroup
  \makeatletter
  \providecommand\color[2][]{%
    \GenericError{(gnuplot) \space\space\space\@spaces}{%
      Package color not loaded in conjunction with
      terminal option `colourtext'%
    }{See the gnuplot documentation for explanation.%
    }{Either use 'blacktext' in gnuplot or load the package
      color.sty in LaTeX.}%
    \renewcommand\color[2][]{}%
  }%
  \providecommand\includegraphics[2][]{%
    \GenericError{(gnuplot) \space\space\space\@spaces}{%
      Package graphicx or graphics not loaded%
    }{See the gnuplot documentation for explanation.%
    }{The gnuplot epslatex terminal needs graphicx.sty or graphics.sty.}%
    \renewcommand\includegraphics[2][]{}%
  }%
  \providecommand\rotatebox[2]{#2}%
  \@ifundefined{ifGPcolor}{%
    \newif\ifGPcolor
    \GPcolortrue
  }{}%
  \@ifundefined{ifGPblacktext}{%
    \newif\ifGPblacktext
    \GPblacktextfalse
  }{}%
  \let\gplgaddtomacro\g@addto@macro
  \gdef\gplbacktext{}%
  \gdef\gplfronttext{}%
  \makeatother
  \ifGPblacktext
    \def\colorrgb#1{}%
    \def\colorgray#1{}%
  \else
    \ifGPcolor
      \def\colorrgb#1{\color[rgb]{#1}}%
      \def\colorgray#1{\color[gray]{#1}}%
      \expandafter\def\csname LTw\endcsname{\color{white}}%
      \expandafter\def\csname LTb\endcsname{\color{black}}%
      \expandafter\def\csname LTa\endcsname{\color{black}}%
      \expandafter\def\csname LT0\endcsname{\color[rgb]{1,0,0}}%
      \expandafter\def\csname LT1\endcsname{\color[rgb]{0,1,0}}%
      \expandafter\def\csname LT2\endcsname{\color[rgb]{0,0,1}}%
      \expandafter\def\csname LT3\endcsname{\color[rgb]{1,0,1}}%
      \expandafter\def\csname LT4\endcsname{\color[rgb]{0,1,1}}%
      \expandafter\def\csname LT5\endcsname{\color[rgb]{1,1,0}}%
      \expandafter\def\csname LT6\endcsname{\color[rgb]{0,0,0}}%
      \expandafter\def\csname LT7\endcsname{\color[rgb]{1,0.3,0}}%
      \expandafter\def\csname LT8\endcsname{\color[rgb]{0.5,0.5,0.5}}%
    \else
      \def\colorrgb#1{\color{black}}%
      \def\colorgray#1{\color[gray]{#1}}%
      \expandafter\def\csname LTw\endcsname{\color{white}}%
      \expandafter\def\csname LTb\endcsname{\color{black}}%
      \expandafter\def\csname LTa\endcsname{\color{black}}%
      \expandafter\def\csname LT0\endcsname{\color{black}}%
      \expandafter\def\csname LT1\endcsname{\color{black}}%
      \expandafter\def\csname LT2\endcsname{\color{black}}%
      \expandafter\def\csname LT3\endcsname{\color{black}}%
      \expandafter\def\csname LT4\endcsname{\color{black}}%
      \expandafter\def\csname LT5\endcsname{\color{black}}%
      \expandafter\def\csname LT6\endcsname{\color{black}}%
      \expandafter\def\csname LT7\endcsname{\color{black}}%
      \expandafter\def\csname LT8\endcsname{\color{black}}%
    \fi
  \fi
    \setlength{\unitlength}{0.0500bp}%
    \ifx\gptboxheight\undefined%
      \newlength{\gptboxheight}%
      \newlength{\gptboxwidth}%
      \newsavebox{\gptboxtext}%
    \fi%
    \setlength{\fboxrule}{0.5pt}%
    \setlength{\fboxsep}{1pt}%
\begin{picture}(3240.00,2016.00)%
    \gplgaddtomacro\gplbacktext{%
    }%
    \gplgaddtomacro\gplfronttext{%
      \csname LTb\endcsname%
      \put(1586,1906){\makebox(0,0){\strut{}Top-$1$: $\Set{10,14,18,24,26,56,58,60,62,64}$}}%
      \csname LTb\endcsname%
      \put(1716,817){\makebox(0,0)[l]{\strut{}}}%
      \csname LTb\endcsname%
      \settowidth{\gptboxwidth}{\widthof{\shortstack{$\mathbf{0.993}$\\$\mathbf{0.039}$}}}
	\advance\gptboxwidth by 2\fboxsep
      \savebox{\gptboxtext}{\parbox[c][\totalheight+2\fboxsep]{\gptboxwidth}{\centering{\shortstack{$\mathbf{0.993}$\\$\mathbf{0.039}$}}}}
	\put(2663,621){\makebox[-\width][c]{\colorbox{white}{\usebox{\gptboxtext}}}}
	\settowidth{\gptboxwidth}{\usebox{\gptboxtext}}
	\advance\gptboxwidth by 2\fboxsep
	\put(2663,621){\makebox[-\width][c]{\framebox[\gptboxwidth][c]{\usebox{\gptboxtext}}}}
    }%
    \gplbacktext
    \put(0,0){\includegraphics{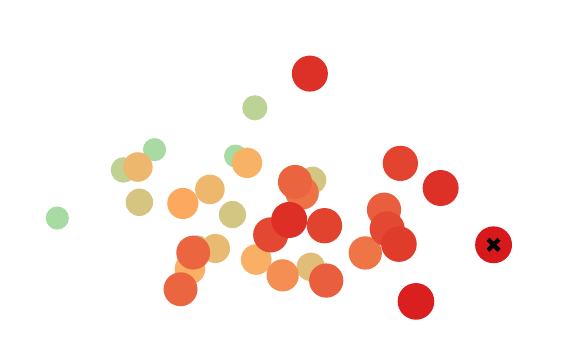}}%
    \gplfronttext
  \end{picture}%
\endgroup

%% file: cells-surp+lymph+48-pca-6.tex
\begingroup
  \makeatletter
  \providecommand\color[2][]{%
    \GenericError{(gnuplot) \space\space\space\@spaces}{%
      Package color not loaded in conjunction with
      terminal option `colourtext'%
    }{See the gnuplot documentation for explanation.%
    }{Either use 'blacktext' in gnuplot or load the package
      color.sty in LaTeX.}%
    \renewcommand\color[2][]{}%
  }%
  \providecommand\includegraphics[2][]{%
    \GenericError{(gnuplot) \space\space\space\@spaces}{%
      Package graphicx or graphics not loaded%
    }{See the gnuplot documentation for explanation.%
    }{The gnuplot epslatex terminal needs graphicx.sty or graphics.sty.}%
    \renewcommand\includegraphics[2][]{}%
  }%
  \providecommand\rotatebox[2]{#2}%
  \@ifundefined{ifGPcolor}{%
    \newif\ifGPcolor
    \GPcolortrue
  }{}%
  \@ifundefined{ifGPblacktext}{%
    \newif\ifGPblacktext
    \GPblacktextfalse
  }{}%
  \let\gplgaddtomacro\g@addto@macro
  \gdef\gplbacktext{}%
  \gdef\gplfronttext{}%
  \makeatother
  \ifGPblacktext
    \def\colorrgb#1{}%
    \def\colorgray#1{}%
  \else
    \ifGPcolor
      \def\colorrgb#1{\color[rgb]{#1}}%
      \def\colorgray#1{\color[gray]{#1}}%
      \expandafter\def\csname LTw\endcsname{\color{white}}%
      \expandafter\def\csname LTb\endcsname{\color{black}}%
      \expandafter\def\csname LTa\endcsname{\color{black}}%
      \expandafter\def\csname LT0\endcsname{\color[rgb]{1,0,0}}%
      \expandafter\def\csname LT1\endcsname{\color[rgb]{0,1,0}}%
      \expandafter\def\csname LT2\endcsname{\color[rgb]{0,0,1}}%
      \expandafter\def\csname LT3\endcsname{\color[rgb]{1,0,1}}%
      \expandafter\def\csname LT4\endcsname{\color[rgb]{0,1,1}}%
      \expandafter\def\csname LT5\endcsname{\color[rgb]{1,1,0}}%
      \expandafter\def\csname LT6\endcsname{\color[rgb]{0,0,0}}%
      \expandafter\def\csname LT7\endcsname{\color[rgb]{1,0.3,0}}%
      \expandafter\def\csname LT8\endcsname{\color[rgb]{0.5,0.5,0.5}}%
    \else
      \def\colorrgb#1{\color{black}}%
      \def\colorgray#1{\color[gray]{#1}}%
      \expandafter\def\csname LTw\endcsname{\color{white}}%
      \expandafter\def\csname LTb\endcsname{\color{black}}%
      \expandafter\def\csname LTa\endcsname{\color{black}}%
      \expandafter\def\csname LT0\endcsname{\color{black}}%
      \expandafter\def\csname LT1\endcsname{\color{black}}%
      \expandafter\def\csname LT2\endcsname{\color{black}}%
      \expandafter\def\csname LT3\endcsname{\color{black}}%
      \expandafter\def\csname LT4\endcsname{\color{black}}%
      \expandafter\def\csname LT5\endcsname{\color{black}}%
      \expandafter\def\csname LT6\endcsname{\color{black}}%
      \expandafter\def\csname LT7\endcsname{\color{black}}%
      \expandafter\def\csname LT8\endcsname{\color{black}}%
    \fi
  \fi
    \setlength{\unitlength}{0.0500bp}%
    \ifx\gptboxheight\undefined%
      \newlength{\gptboxheight}%
      \newlength{\gptboxwidth}%
      \newsavebox{\gptboxtext}%
    \fi%
    \setlength{\fboxrule}{0.5pt}%
    \setlength{\fboxsep}{1pt}%
\begin{picture}(3240.00,2016.00)%
    \gplgaddtomacro\gplbacktext{%
    }%
    \gplgaddtomacro\gplfronttext{%
      \csname LTb\endcsname%
      \put(1586,1906){\makebox(0,0){\strut{}Top-$1$: $\Set{6,8,10,20,24,26,56,60,64,66}$}}%
      \csname LTb\endcsname%
      \put(1590,829){\makebox(0,0)[l]{\strut{}}}%
      \csname LTb\endcsname%
      \settowidth{\gptboxwidth}{\widthof{\shortstack{$\mathbf{0.980}$\\$\mathbf{0.036}$}}}
	\advance\gptboxwidth by 2\fboxsep
      \savebox{\gptboxtext}{\parbox[c][\totalheight+2\fboxsep]{\gptboxwidth}{\centering{\shortstack{$\mathbf{0.980}$\\$\mathbf{0.036}$}}}}
	\put(2231,1571){\makebox[-\width][c]{\colorbox{white}{\usebox{\gptboxtext}}}}
	\settowidth{\gptboxwidth}{\usebox{\gptboxtext}}
	\advance\gptboxwidth by 2\fboxsep
	\put(2231,1571){\makebox[-\width][c]{\framebox[\gptboxwidth][c]{\usebox{\gptboxtext}}}}
    }%
    \gplbacktext
    \put(0,0){\includegraphics{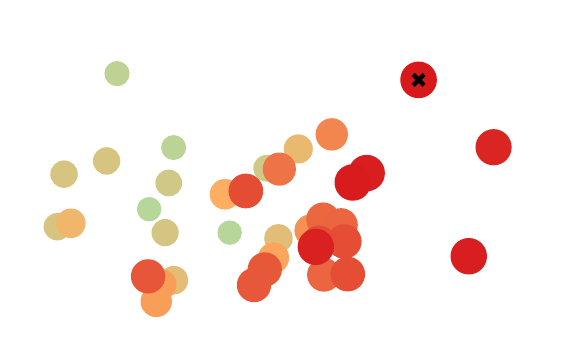}}%
    \gplfronttext
  \end{picture}%
\endgroup

%% file: cells-surp+lymph+48-pca-4.tex
\begingroup
  \makeatletter
  \providecommand\color[2][]{%
    \GenericError{(gnuplot) \space\space\space\@spaces}{%
      Package color not loaded in conjunction with
      terminal option `colourtext'%
    }{See the gnuplot documentation for explanation.%
    }{Either use 'blacktext' in gnuplot or load the package
      color.sty in LaTeX.}%
    \renewcommand\color[2][]{}%
  }%
  \providecommand\includegraphics[2][]{%
    \GenericError{(gnuplot) \space\space\space\@spaces}{%
      Package graphicx or graphics not loaded%
    }{See the gnuplot documentation for explanation.%
    }{The gnuplot epslatex terminal needs graphicx.sty or graphics.sty.}%
    \renewcommand\includegraphics[2][]{}%
  }%
  \providecommand\rotatebox[2]{#2}%
  \@ifundefined{ifGPcolor}{%
    \newif\ifGPcolor
    \GPcolortrue
  }{}%
  \@ifundefined{ifGPblacktext}{%
    \newif\ifGPblacktext
    \GPblacktextfalse
  }{}%
  \let\gplgaddtomacro\g@addto@macro
  \gdef\gplbacktext{}%
  \gdef\gplfronttext{}%
  \makeatother
  \ifGPblacktext
    \def\colorrgb#1{}%
    \def\colorgray#1{}%
  \else
    \ifGPcolor
      \def\colorrgb#1{\color[rgb]{#1}}%
      \def\colorgray#1{\color[gray]{#1}}%
      \expandafter\def\csname LTw\endcsname{\color{white}}%
      \expandafter\def\csname LTb\endcsname{\color{black}}%
      \expandafter\def\csname LTa\endcsname{\color{black}}%
      \expandafter\def\csname LT0\endcsname{\color[rgb]{1,0,0}}%
      \expandafter\def\csname LT1\endcsname{\color[rgb]{0,1,0}}%
      \expandafter\def\csname LT2\endcsname{\color[rgb]{0,0,1}}%
      \expandafter\def\csname LT3\endcsname{\color[rgb]{1,0,1}}%
      \expandafter\def\csname LT4\endcsname{\color[rgb]{0,1,1}}%
      \expandafter\def\csname LT5\endcsname{\color[rgb]{1,1,0}}%
      \expandafter\def\csname LT6\endcsname{\color[rgb]{0,0,0}}%
      \expandafter\def\csname LT7\endcsname{\color[rgb]{1,0.3,0}}%
      \expandafter\def\csname LT8\endcsname{\color[rgb]{0.5,0.5,0.5}}%
    \else
      \def\colorrgb#1{\color{black}}%
      \def\colorgray#1{\color[gray]{#1}}%
      \expandafter\def\csname LTw\endcsname{\color{white}}%
      \expandafter\def\csname LTb\endcsname{\color{black}}%
      \expandafter\def\csname LTa\endcsname{\color{black}}%
      \expandafter\def\csname LT0\endcsname{\color{black}}%
      \expandafter\def\csname LT1\endcsname{\color{black}}%
      \expandafter\def\csname LT2\endcsname{\color{black}}%
      \expandafter\def\csname LT3\endcsname{\color{black}}%
      \expandafter\def\csname LT4\endcsname{\color{black}}%
      \expandafter\def\csname LT5\endcsname{\color{black}}%
      \expandafter\def\csname LT6\endcsname{\color{black}}%
      \expandafter\def\csname LT7\endcsname{\color{black}}%
      \expandafter\def\csname LT8\endcsname{\color{black}}%
    \fi
  \fi
    \setlength{\unitlength}{0.0500bp}%
    \ifx\gptboxheight\undefined%
      \newlength{\gptboxheight}%
      \newlength{\gptboxwidth}%
      \newsavebox{\gptboxtext}%
    \fi%
    \setlength{\fboxrule}{0.5pt}%
    \setlength{\fboxsep}{1pt}%
\begin{picture}(3240.00,2016.00)%
    \gplgaddtomacro\gplbacktext{%
    }%
    \gplgaddtomacro\gplfronttext{%
      \csname LTb\endcsname%
      \put(1586,1906){\makebox(0,0){\strut{}Top-$1$: $\Set{6,10,14,18,23,24,26,56,58,60,64,66}$}}%
      \csname LTb\endcsname%
      \put(1833,926){\makebox(0,0)[l]{\strut{}}}%
      \csname LTb\endcsname%
      \settowidth{\gptboxwidth}{\widthof{\shortstack{$\mathbf{0.985}$\\$\mathbf{0.034}$}}}
	\advance\gptboxwidth by 2\fboxsep
      \savebox{\gptboxtext}{\parbox[c][\totalheight+2\fboxsep]{\gptboxwidth}{\centering{\shortstack{$\mathbf{0.985}$\\$\mathbf{0.034}$}}}}
	\put(2663,1077){\makebox[-\width][c]{\colorbox{white}{\usebox{\gptboxtext}}}}
	\settowidth{\gptboxwidth}{\usebox{\gptboxtext}}
	\advance\gptboxwidth by 2\fboxsep
	\put(2663,1077){\makebox[-\width][c]{\framebox[\gptboxwidth][c]{\usebox{\gptboxtext}}}}
    }%
    \gplbacktext
    \put(0,0){\includegraphics{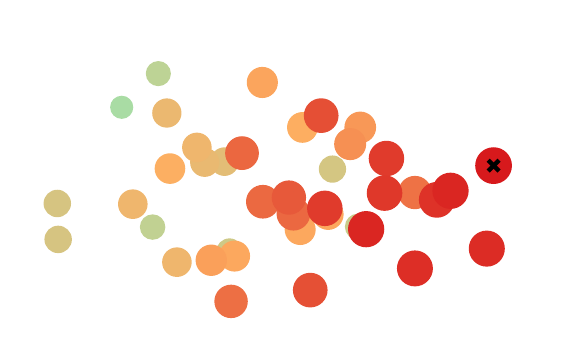}}%
    \gplfronttext
  \end{picture}%
\endgroup

%% file: cells-surp+lymph+48-pca-5.tex
\begingroup
  \makeatletter
  \providecommand\color[2][]{%
    \GenericError{(gnuplot) \space\space\space\@spaces}{%
      Package color not loaded in conjunction with
      terminal option `colourtext'%
    }{See the gnuplot documentation for explanation.%
    }{Either use 'blacktext' in gnuplot or load the package
      color.sty in LaTeX.}%
    \renewcommand\color[2][]{}%
  }%
  \providecommand\includegraphics[2][]{%
    \GenericError{(gnuplot) \space\space\space\@spaces}{%
      Package graphicx or graphics not loaded%
    }{See the gnuplot documentation for explanation.%
    }{The gnuplot epslatex terminal needs graphicx.sty or graphics.sty.}%
    \renewcommand\includegraphics[2][]{}%
  }%
  \providecommand\rotatebox[2]{#2}%
  \@ifundefined{ifGPcolor}{%
    \newif\ifGPcolor
    \GPcolortrue
  }{}%
  \@ifundefined{ifGPblacktext}{%
    \newif\ifGPblacktext
    \GPblacktextfalse
  }{}%
  \let\gplgaddtomacro\g@addto@macro
  \gdef\gplbacktext{}%
  \gdef\gplfronttext{}%
  \makeatother
  \ifGPblacktext
    \def\colorrgb#1{}%
    \def\colorgray#1{}%
  \else
    \ifGPcolor
      \def\colorrgb#1{\color[rgb]{#1}}%
      \def\colorgray#1{\color[gray]{#1}}%
      \expandafter\def\csname LTw\endcsname{\color{white}}%
      \expandafter\def\csname LTb\endcsname{\color{black}}%
      \expandafter\def\csname LTa\endcsname{\color{black}}%
      \expandafter\def\csname LT0\endcsname{\color[rgb]{1,0,0}}%
      \expandafter\def\csname LT1\endcsname{\color[rgb]{0,1,0}}%
      \expandafter\def\csname LT2\endcsname{\color[rgb]{0,0,1}}%
      \expandafter\def\csname LT3\endcsname{\color[rgb]{1,0,1}}%
      \expandafter\def\csname LT4\endcsname{\color[rgb]{0,1,1}}%
      \expandafter\def\csname LT5\endcsname{\color[rgb]{1,1,0}}%
      \expandafter\def\csname LT6\endcsname{\color[rgb]{0,0,0}}%
      \expandafter\def\csname LT7\endcsname{\color[rgb]{1,0.3,0}}%
      \expandafter\def\csname LT8\endcsname{\color[rgb]{0.5,0.5,0.5}}%
    \else
      \def\colorrgb#1{\color{black}}%
      \def\colorgray#1{\color[gray]{#1}}%
      \expandafter\def\csname LTw\endcsname{\color{white}}%
      \expandafter\def\csname LTb\endcsname{\color{black}}%
      \expandafter\def\csname LTa\endcsname{\color{black}}%
      \expandafter\def\csname LT0\endcsname{\color{black}}%
      \expandafter\def\csname LT1\endcsname{\color{black}}%
      \expandafter\def\csname LT2\endcsname{\color{black}}%
      \expandafter\def\csname LT3\endcsname{\color{black}}%
      \expandafter\def\csname LT4\endcsname{\color{black}}%
      \expandafter\def\csname LT5\endcsname{\color{black}}%
      \expandafter\def\csname LT6\endcsname{\color{black}}%
      \expandafter\def\csname LT7\endcsname{\color{black}}%
      \expandafter\def\csname LT8\endcsname{\color{black}}%
    \fi
  \fi
    \setlength{\unitlength}{0.0500bp}%
    \ifx\gptboxheight\undefined%
      \newlength{\gptboxheight}%
      \newlength{\gptboxwidth}%
      \newsavebox{\gptboxtext}%
    \fi%
    \setlength{\fboxrule}{0.5pt}%
    \setlength{\fboxsep}{1pt}%
\begin{picture}(3240.00,2016.00)%
    \gplgaddtomacro\gplbacktext{%
    }%
    \gplgaddtomacro\gplfronttext{%
      \csname LTb\endcsname%
      \put(1586,1906){\makebox(0,0){\strut{}Top-$1$: $\Set{8,20,24,26,56,60,62}$}}%
      \csname LTb\endcsname%
      \put(1924,943){\makebox(0,0)[l]{\strut{}}}%
      \csname LTb\endcsname%
      \settowidth{\gptboxwidth}{\widthof{\shortstack{$\mathbf{0.980}$\\$\mathbf{0.038}$}}}
	\advance\gptboxwidth by 2\fboxsep
      \savebox{\gptboxtext}{\parbox[c][\totalheight+2\fboxsep]{\gptboxwidth}{\centering{\shortstack{$\mathbf{0.980}$\\$\mathbf{0.038}$}}}}
	\put(2397,764){\makebox[-\width][c]{\colorbox{white}{\usebox{\gptboxtext}}}}
	\settowidth{\gptboxwidth}{\usebox{\gptboxtext}}
	\advance\gptboxwidth by 2\fboxsep
	\put(2397,764){\makebox[-\width][c]{\framebox[\gptboxwidth][c]{\usebox{\gptboxtext}}}}
    }%
    \gplbacktext
    \put(0,0){\includegraphics{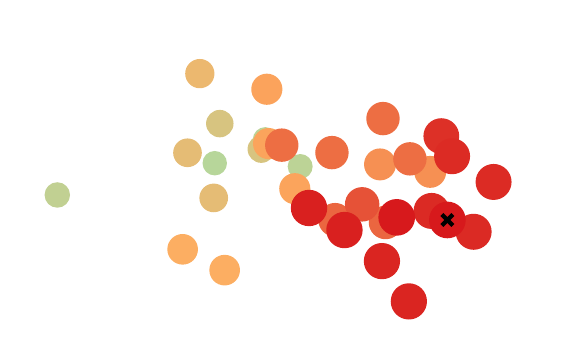}}%
    \gplfronttext
  \end{picture}%
\endgroup

%% file: cells-surp+lymph+48-pca-7.tex
\begingroup
  \makeatletter
  \providecommand\color[2][]{%
    \GenericError{(gnuplot) \space\space\space\@spaces}{%
      Package color not loaded in conjunction with
      terminal option `colourtext'%
    }{See the gnuplot documentation for explanation.%
    }{Either use 'blacktext' in gnuplot or load the package
      color.sty in LaTeX.}%
    \renewcommand\color[2][]{}%
  }%
  \providecommand\includegraphics[2][]{%
    \GenericError{(gnuplot) \space\space\space\@spaces}{%
      Package graphicx or graphics not loaded%
    }{See the gnuplot documentation for explanation.%
    }{The gnuplot epslatex terminal needs graphicx.sty or graphics.sty.}%
    \renewcommand\includegraphics[2][]{}%
  }%
  \providecommand\rotatebox[2]{#2}%
  \@ifundefined{ifGPcolor}{%
    \newif\ifGPcolor
    \GPcolortrue
  }{}%
  \@ifundefined{ifGPblacktext}{%
    \newif\ifGPblacktext
    \GPblacktextfalse
  }{}%
  \let\gplgaddtomacro\g@addto@macro
  \gdef\gplbacktext{}%
  \gdef\gplfronttext{}%
  \makeatother
  \ifGPblacktext
    \def\colorrgb#1{}%
    \def\colorgray#1{}%
  \else
    \ifGPcolor
      \def\colorrgb#1{\color[rgb]{#1}}%
      \def\colorgray#1{\color[gray]{#1}}%
      \expandafter\def\csname LTw\endcsname{\color{white}}%
      \expandafter\def\csname LTb\endcsname{\color{black}}%
      \expandafter\def\csname LTa\endcsname{\color{black}}%
      \expandafter\def\csname LT0\endcsname{\color[rgb]{1,0,0}}%
      \expandafter\def\csname LT1\endcsname{\color[rgb]{0,1,0}}%
      \expandafter\def\csname LT2\endcsname{\color[rgb]{0,0,1}}%
      \expandafter\def\csname LT3\endcsname{\color[rgb]{1,0,1}}%
      \expandafter\def\csname LT4\endcsname{\color[rgb]{0,1,1}}%
      \expandafter\def\csname LT5\endcsname{\color[rgb]{1,1,0}}%
      \expandafter\def\csname LT6\endcsname{\color[rgb]{0,0,0}}%
      \expandafter\def\csname LT7\endcsname{\color[rgb]{1,0.3,0}}%
      \expandafter\def\csname LT8\endcsname{\color[rgb]{0.5,0.5,0.5}}%
    \else
      \def\colorrgb#1{\color{black}}%
      \def\colorgray#1{\color[gray]{#1}}%
      \expandafter\def\csname LTw\endcsname{\color{white}}%
      \expandafter\def\csname LTb\endcsname{\color{black}}%
      \expandafter\def\csname LTa\endcsname{\color{black}}%
      \expandafter\def\csname LT0\endcsname{\color{black}}%
      \expandafter\def\csname LT1\endcsname{\color{black}}%
      \expandafter\def\csname LT2\endcsname{\color{black}}%
      \expandafter\def\csname LT3\endcsname{\color{black}}%
      \expandafter\def\csname LT4\endcsname{\color{black}}%
      \expandafter\def\csname LT5\endcsname{\color{black}}%
      \expandafter\def\csname LT6\endcsname{\color{black}}%
      \expandafter\def\csname LT7\endcsname{\color{black}}%
      \expandafter\def\csname LT8\endcsname{\color{black}}%
    \fi
  \fi
    \setlength{\unitlength}{0.0500bp}%
    \ifx\gptboxheight\undefined%
      \newlength{\gptboxheight}%
      \newlength{\gptboxwidth}%
      \newsavebox{\gptboxtext}%
    \fi%
    \setlength{\fboxrule}{0.5pt}%
    \setlength{\fboxsep}{1pt}%
\begin{picture}(3240.00,2016.00)%
    \gplgaddtomacro\gplbacktext{%
    }%
    \gplgaddtomacro\gplfronttext{%
      \csname LTb\endcsname%
      \put(1586,1906){\makebox(0,0){\strut{}Top-$1$: $\Set{6,8,10,20,24,26,56,58,60,64}$}}%
      \csname LTb\endcsname%
      \put(1515,807){\makebox(0,0)[l]{\strut{}}}%
      \csname LTb\endcsname%
      \settowidth{\gptboxwidth}{\widthof{\shortstack{$\mathbf{0.993}$\\$\mathbf{0.041}$}}}
	\advance\gptboxwidth by 2\fboxsep
      \savebox{\gptboxtext}{\parbox[c][\totalheight+2\fboxsep]{\gptboxwidth}{\centering{\shortstack{$\mathbf{0.993}$\\$\mathbf{0.041}$}}}}
	\put(2663,1299){\makebox[-\width][c]{\colorbox{white}{\usebox{\gptboxtext}}}}
	\settowidth{\gptboxwidth}{\usebox{\gptboxtext}}
	\advance\gptboxwidth by 2\fboxsep
	\put(2663,1299){\makebox[-\width][c]{\framebox[\gptboxwidth][c]{\usebox{\gptboxtext}}}}
    }%
    \gplbacktext
    \put(0,0){\includegraphics{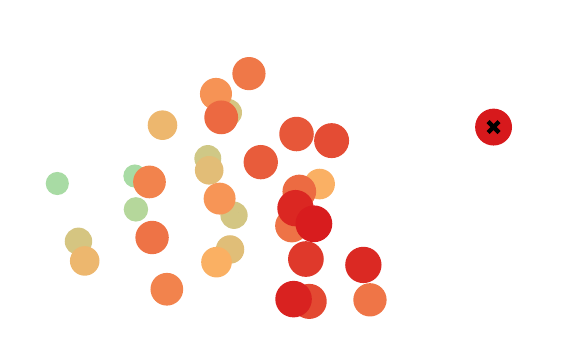}}%
    \gplfronttext
  \end{picture}%
\endgroup

%% file: cells-surp+lymph+48-pca-8.tex
\begingroup
  \makeatletter
  \providecommand\color[2][]{%
    \GenericError{(gnuplot) \space\space\space\@spaces}{%
      Package color not loaded in conjunction with
      terminal option `colourtext'%
    }{See the gnuplot documentation for explanation.%
    }{Either use 'blacktext' in gnuplot or load the package
      color.sty in LaTeX.}%
    \renewcommand\color[2][]{}%
  }%
  \providecommand\includegraphics[2][]{%
    \GenericError{(gnuplot) \space\space\space\@spaces}{%
      Package graphicx or graphics not loaded%
    }{See the gnuplot documentation for explanation.%
    }{The gnuplot epslatex terminal needs graphicx.sty or graphics.sty.}%
    \renewcommand\includegraphics[2][]{}%
  }%
  \providecommand\rotatebox[2]{#2}%
  \@ifundefined{ifGPcolor}{%
    \newif\ifGPcolor
    \GPcolortrue
  }{}%
  \@ifundefined{ifGPblacktext}{%
    \newif\ifGPblacktext
    \GPblacktextfalse
  }{}%
  \let\gplgaddtomacro\g@addto@macro
  \gdef\gplbacktext{}%
  \gdef\gplfronttext{}%
  \makeatother
  \ifGPblacktext
    \def\colorrgb#1{}%
    \def\colorgray#1{}%
  \else
    \ifGPcolor
      \def\colorrgb#1{\color[rgb]{#1}}%
      \def\colorgray#1{\color[gray]{#1}}%
      \expandafter\def\csname LTw\endcsname{\color{white}}%
      \expandafter\def\csname LTb\endcsname{\color{black}}%
      \expandafter\def\csname LTa\endcsname{\color{black}}%
      \expandafter\def\csname LT0\endcsname{\color[rgb]{1,0,0}}%
      \expandafter\def\csname LT1\endcsname{\color[rgb]{0,1,0}}%
      \expandafter\def\csname LT2\endcsname{\color[rgb]{0,0,1}}%
      \expandafter\def\csname LT3\endcsname{\color[rgb]{1,0,1}}%
      \expandafter\def\csname LT4\endcsname{\color[rgb]{0,1,1}}%
      \expandafter\def\csname LT5\endcsname{\color[rgb]{1,1,0}}%
      \expandafter\def\csname LT6\endcsname{\color[rgb]{0,0,0}}%
      \expandafter\def\csname LT7\endcsname{\color[rgb]{1,0.3,0}}%
      \expandafter\def\csname LT8\endcsname{\color[rgb]{0.5,0.5,0.5}}%
    \else
      \def\colorrgb#1{\color{black}}%
      \def\colorgray#1{\color[gray]{#1}}%
      \expandafter\def\csname LTw\endcsname{\color{white}}%
      \expandafter\def\csname LTb\endcsname{\color{black}}%
      \expandafter\def\csname LTa\endcsname{\color{black}}%
      \expandafter\def\csname LT0\endcsname{\color{black}}%
      \expandafter\def\csname LT1\endcsname{\color{black}}%
      \expandafter\def\csname LT2\endcsname{\color{black}}%
      \expandafter\def\csname LT3\endcsname{\color{black}}%
      \expandafter\def\csname LT4\endcsname{\color{black}}%
      \expandafter\def\csname LT5\endcsname{\color{black}}%
      \expandafter\def\csname LT6\endcsname{\color{black}}%
      \expandafter\def\csname LT7\endcsname{\color{black}}%
      \expandafter\def\csname LT8\endcsname{\color{black}}%
    \fi
  \fi
    \setlength{\unitlength}{0.0500bp}%
    \ifx\gptboxheight\undefined%
      \newlength{\gptboxheight}%
      \newlength{\gptboxwidth}%
      \newsavebox{\gptboxtext}%
    \fi%
    \setlength{\fboxrule}{0.5pt}%
    \setlength{\fboxsep}{1pt}%
\begin{picture}(3240.00,2016.00)%
    \gplgaddtomacro\gplbacktext{%
    }%
    \gplgaddtomacro\gplfronttext{%
      \csname LTb\endcsname%
      \put(1586,1906){\makebox(0,0){\strut{}Top-$1$: $\Set{14,18,20,23,37,56,60,62,64}$}}%
      \csname LTb\endcsname%
      \put(1901,985){\makebox(0,0)[l]{\strut{}}}%
      \csname LTb\endcsname%
      \settowidth{\gptboxwidth}{\widthof{\shortstack{$\mathbf{0.965}$\\$\mathbf{0.032}$}}}
	\advance\gptboxwidth by 2\fboxsep
      \savebox{\gptboxtext}{\parbox[c][\totalheight+2\fboxsep]{\gptboxwidth}{\centering{\shortstack{$\mathbf{0.965}$\\$\mathbf{0.032}$}}}}
	\put(1611,295){\makebox[-\width][c]{\colorbox{white}{\usebox{\gptboxtext}}}}
	\settowidth{\gptboxwidth}{\usebox{\gptboxtext}}
	\advance\gptboxwidth by 2\fboxsep
	\put(1611,295){\makebox[-\width][c]{\framebox[\gptboxwidth][c]{\usebox{\gptboxtext}}}}
    }%
    \gplbacktext
    \put(0,0){\includegraphics{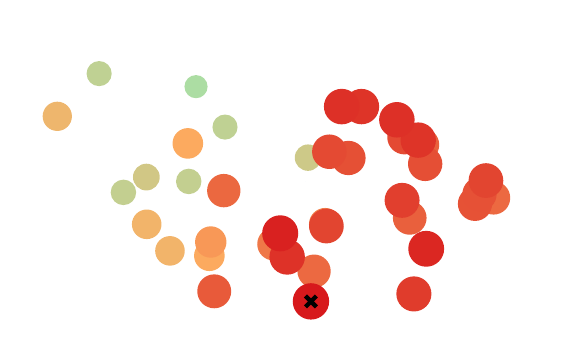}}%
    \gplfronttext
  \end{picture}%
\endgroup

%% file: cells-surp+lymph+48-pca-9.tex
\begingroup
  \makeatletter
  \providecommand\color[2][]{%
    \GenericError{(gnuplot) \space\space\space\@spaces}{%
      Package color not loaded in conjunction with
      terminal option `colourtext'%
    }{See the gnuplot documentation for explanation.%
    }{Either use 'blacktext' in gnuplot or load the package
      color.sty in LaTeX.}%
    \renewcommand\color[2][]{}%
  }%
  \providecommand\includegraphics[2][]{%
    \GenericError{(gnuplot) \space\space\space\@spaces}{%
      Package graphicx or graphics not loaded%
    }{See the gnuplot documentation for explanation.%
    }{The gnuplot epslatex terminal needs graphicx.sty or graphics.sty.}%
    \renewcommand\includegraphics[2][]{}%
  }%
  \providecommand\rotatebox[2]{#2}%
  \@ifundefined{ifGPcolor}{%
    \newif\ifGPcolor
    \GPcolortrue
  }{}%
  \@ifundefined{ifGPblacktext}{%
    \newif\ifGPblacktext
    \GPblacktextfalse
  }{}%
  \let\gplgaddtomacro\g@addto@macro
  \gdef\gplbacktext{}%
  \gdef\gplfronttext{}%
  \makeatother
  \ifGPblacktext
    \def\colorrgb#1{}%
    \def\colorgray#1{}%
  \else
    \ifGPcolor
      \def\colorrgb#1{\color[rgb]{#1}}%
      \def\colorgray#1{\color[gray]{#1}}%
      \expandafter\def\csname LTw\endcsname{\color{white}}%
      \expandafter\def\csname LTb\endcsname{\color{black}}%
      \expandafter\def\csname LTa\endcsname{\color{black}}%
      \expandafter\def\csname LT0\endcsname{\color[rgb]{1,0,0}}%
      \expandafter\def\csname LT1\endcsname{\color[rgb]{0,1,0}}%
      \expandafter\def\csname LT2\endcsname{\color[rgb]{0,0,1}}%
      \expandafter\def\csname LT3\endcsname{\color[rgb]{1,0,1}}%
      \expandafter\def\csname LT4\endcsname{\color[rgb]{0,1,1}}%
      \expandafter\def\csname LT5\endcsname{\color[rgb]{1,1,0}}%
      \expandafter\def\csname LT6\endcsname{\color[rgb]{0,0,0}}%
      \expandafter\def\csname LT7\endcsname{\color[rgb]{1,0.3,0}}%
      \expandafter\def\csname LT8\endcsname{\color[rgb]{0.5,0.5,0.5}}%
    \else
      \def\colorrgb#1{\color{black}}%
      \def\colorgray#1{\color[gray]{#1}}%
      \expandafter\def\csname LTw\endcsname{\color{white}}%
      \expandafter\def\csname LTb\endcsname{\color{black}}%
      \expandafter\def\csname LTa\endcsname{\color{black}}%
      \expandafter\def\csname LT0\endcsname{\color{black}}%
      \expandafter\def\csname LT1\endcsname{\color{black}}%
      \expandafter\def\csname LT2\endcsname{\color{black}}%
      \expandafter\def\csname LT3\endcsname{\color{black}}%
      \expandafter\def\csname LT4\endcsname{\color{black}}%
      \expandafter\def\csname LT5\endcsname{\color{black}}%
      \expandafter\def\csname LT6\endcsname{\color{black}}%
      \expandafter\def\csname LT7\endcsname{\color{black}}%
      \expandafter\def\csname LT8\endcsname{\color{black}}%
    \fi
  \fi
    \setlength{\unitlength}{0.0500bp}%
    \ifx\gptboxheight\undefined%
      \newlength{\gptboxheight}%
      \newlength{\gptboxwidth}%
      \newsavebox{\gptboxtext}%
    \fi%
    \setlength{\fboxrule}{0.5pt}%
    \setlength{\fboxsep}{1pt}%
\begin{picture}(3240.00,2016.00)%
    \gplgaddtomacro\gplbacktext{%
    }%
    \gplgaddtomacro\gplfronttext{%
      \csname LTb\endcsname%
      \put(1586,1906){\makebox(0,0){\strut{}Top-$1$: $\Set{8,20,23,26,29,56,60,64}$}}%
      \csname LTb\endcsname%
      \put(1678,896){\makebox(0,0)[l]{\strut{}}}%
      \csname LTb\endcsname%
      \settowidth{\gptboxwidth}{\widthof{\shortstack{$\mathbf{0.972}$\\$\mathbf{0.036}$}}}
	\advance\gptboxwidth by 2\fboxsep
      \savebox{\gptboxtext}{\parbox[c][\totalheight+2\fboxsep]{\gptboxwidth}{\centering{\shortstack{$\mathbf{0.972}$\\$\mathbf{0.036}$}}}}
	\put(1625,558){\makebox[-\width][c]{\colorbox{white}{\usebox{\gptboxtext}}}}
	\settowidth{\gptboxwidth}{\usebox{\gptboxtext}}
	\advance\gptboxwidth by 2\fboxsep
	\put(1625,558){\makebox[-\width][c]{\framebox[\gptboxwidth][c]{\usebox{\gptboxtext}}}}
    }%
    \gplbacktext
    \put(0,0){\includegraphics{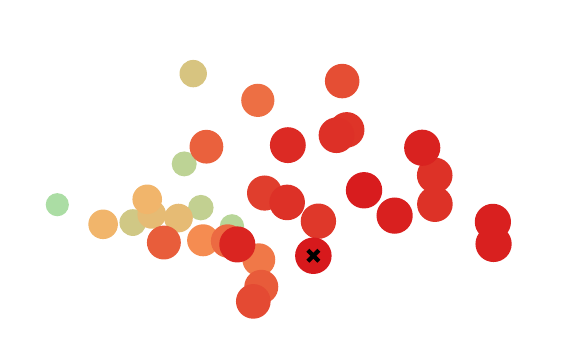}}%
    \gplfronttext
  \end{picture}%
\endgroup

%% file: cells-surp+lymph+48-pca-10.tex
\begingroup
  \makeatletter
  \providecommand\color[2][]{%
    \GenericError{(gnuplot) \space\space\space\@spaces}{%
      Package color not loaded in conjunction with
      terminal option `colourtext'%
    }{See the gnuplot documentation for explanation.%
    }{Either use 'blacktext' in gnuplot or load the package
      color.sty in LaTeX.}%
    \renewcommand\color[2][]{}%
  }%
  \providecommand\includegraphics[2][]{%
    \GenericError{(gnuplot) \space\space\space\@spaces}{%
      Package graphicx or graphics not loaded%
    }{See the gnuplot documentation for explanation.%
    }{The gnuplot epslatex terminal needs graphicx.sty or graphics.sty.}%
    \renewcommand\includegraphics[2][]{}%
  }%
  \providecommand\rotatebox[2]{#2}%
  \@ifundefined{ifGPcolor}{%
    \newif\ifGPcolor
    \GPcolortrue
  }{}%
  \@ifundefined{ifGPblacktext}{%
    \newif\ifGPblacktext
    \GPblacktextfalse
  }{}%
  \let\gplgaddtomacro\g@addto@macro
  \gdef\gplbacktext{}%
  \gdef\gplfronttext{}%
  \makeatother
  \ifGPblacktext
    \def\colorrgb#1{}%
    \def\colorgray#1{}%
  \else
    \ifGPcolor
      \def\colorrgb#1{\color[rgb]{#1}}%
      \def\colorgray#1{\color[gray]{#1}}%
      \expandafter\def\csname LTw\endcsname{\color{white}}%
      \expandafter\def\csname LTb\endcsname{\color{black}}%
      \expandafter\def\csname LTa\endcsname{\color{black}}%
      \expandafter\def\csname LT0\endcsname{\color[rgb]{1,0,0}}%
      \expandafter\def\csname LT1\endcsname{\color[rgb]{0,1,0}}%
      \expandafter\def\csname LT2\endcsname{\color[rgb]{0,0,1}}%
      \expandafter\def\csname LT3\endcsname{\color[rgb]{1,0,1}}%
      \expandafter\def\csname LT4\endcsname{\color[rgb]{0,1,1}}%
      \expandafter\def\csname LT5\endcsname{\color[rgb]{1,1,0}}%
      \expandafter\def\csname LT6\endcsname{\color[rgb]{0,0,0}}%
      \expandafter\def\csname LT7\endcsname{\color[rgb]{1,0.3,0}}%
      \expandafter\def\csname LT8\endcsname{\color[rgb]{0.5,0.5,0.5}}%
    \else
      \def\colorrgb#1{\color{black}}%
      \def\colorgray#1{\color[gray]{#1}}%
      \expandafter\def\csname LTw\endcsname{\color{white}}%
      \expandafter\def\csname LTb\endcsname{\color{black}}%
      \expandafter\def\csname LTa\endcsname{\color{black}}%
      \expandafter\def\csname LT0\endcsname{\color{black}}%
      \expandafter\def\csname LT1\endcsname{\color{black}}%
      \expandafter\def\csname LT2\endcsname{\color{black}}%
      \expandafter\def\csname LT3\endcsname{\color{black}}%
      \expandafter\def\csname LT4\endcsname{\color{black}}%
      \expandafter\def\csname LT5\endcsname{\color{black}}%
      \expandafter\def\csname LT6\endcsname{\color{black}}%
      \expandafter\def\csname LT7\endcsname{\color{black}}%
      \expandafter\def\csname LT8\endcsname{\color{black}}%
    \fi
  \fi
    \setlength{\unitlength}{0.0500bp}%
    \ifx\gptboxheight\undefined%
      \newlength{\gptboxheight}%
      \newlength{\gptboxwidth}%
      \newsavebox{\gptboxtext}%
    \fi%
    \setlength{\fboxrule}{0.5pt}%
    \setlength{\fboxsep}{1pt}%
\begin{picture}(3240.00,2016.00)%
    \gplgaddtomacro\gplbacktext{%
    }%
    \gplgaddtomacro\gplfronttext{%
      \csname LTb\endcsname%
      \put(1586,1906){\makebox(0,0){\strut{}Top-$1$: $\Set{6,10,14,20,24,56,60,62,64}$}}%
      \csname LTb\endcsname%
      \put(1946,854){\makebox(0,0)[l]{\strut{}}}%
      \csname LTb\endcsname%
      \settowidth{\gptboxwidth}{\widthof{\shortstack{$\mathbf{0.988}$\\$\mathbf{0.033}$}}}
	\advance\gptboxwidth by 2\fboxsep
      \savebox{\gptboxtext}{\parbox[c][\totalheight+2\fboxsep]{\gptboxwidth}{\centering{\shortstack{$\mathbf{0.988}$\\$\mathbf{0.033}$}}}}
	\put(2572,1278){\makebox[-\width][c]{\colorbox{white}{\usebox{\gptboxtext}}}}
	\settowidth{\gptboxwidth}{\usebox{\gptboxtext}}
	\advance\gptboxwidth by 2\fboxsep
	\put(2572,1278){\makebox[-\width][c]{\framebox[\gptboxwidth][c]{\usebox{\gptboxtext}}}}
    }%
    \gplbacktext
    \put(0,0){\includegraphics{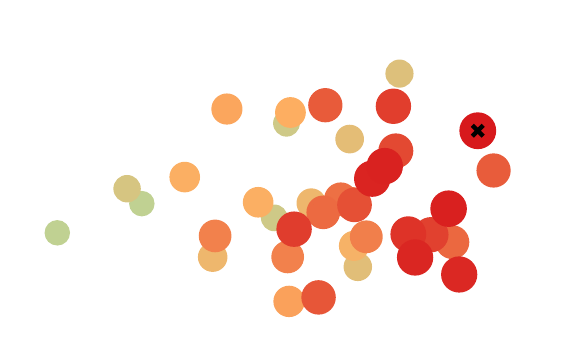}}%
    \gplfronttext
  \end{picture}%
\endgroup